\RequirePackage{fix-cm}
\documentclass{svjour3}                     
\smartqed  
\usepackage{natbib}
\usepackage{mathptmx}      
\usepackage[cmex10]{amsmath}
\usepackage{amsfonts}
\usepackage{amssymb}

\usepackage[pdftex]{graphicx}
\usepackage{epstopdf}
\DeclareGraphicsExtensions{.eps,.png,.jpg,.pdf,.tif,.tiff}
\RequirePackage{epsfig}
\usepackage{subfig}
\graphicspath{{img/raster/TIFF/}{img/vector/}{./}{img/raster/}}

\usepackage{booktabs}
\usepackage{chngpage}
\usepackage{rotating}
\usepackage{multirow}

\RequirePackage{color}
\RequirePackage{xspace}
\RequirePackage{eso-pic}
\usepackage{ucs}
\usepackage[utf8x]{inputenc}	
\usepackage{url}
\usepackage{hyperref}

\usepackage{algorithm}
\usepackage{algorithmic}

\begin{document}

\title{Unsupervised Feature Construction for Improving Data Representation and Semantics}
\titlerunning{Unsupervised Feature Construction}        

\author {
		Marian-Andrei Rizoiu\and%
        Julien Velcin\and\\%
        St\'{e}phane Lallich \\%
}


\institute{Marian-Andrei Rizoiu \and Julien Velcin \and St\'ephane Lallich \at
              ERIC Laboratory, University Lumi\`ere Lyon 2 \\
              5, avenue Pierre Mend\`es France, 69676 Bron Cedex, France \\
              Tel. +33 (0)4 78 77 31 54\\
              Fax. +33 (0)4 78 77 23 75\\
		   \and 
		   Marian-Andrei Rizoiu \at
		   		\email{Marian-Andrei.Rizoiu@univ-lyon2.fr}
           \and
           Julien Velcin \at
              \email{Julien.Velcin@univ-lyon2.fr}
           \and
           St\'ephane Lallich \at
              \email{Stephane.Lallich@univ-lyon2.fr}
}

\date{Received: 27/01/2012 / Accepted: 29/01/2013}

\maketitle

\begin{abstract}
\textcolor{black}{Feature}-based format is the main data representation format used by machine learning algorithms.
When the \textcolor{black}{features} do not properly describe the initial data, performance starts to degrade.
Some algorithms address this problem by internally changing the representation space, but the newly-constructed features are rarely \textcolor{black}{comprehensible}.
We seek to construct, in an unsupervised way, new \textcolor{black}{features} that are more appropriate for describing a given dataset and, at the same time, comprehensible for a human user.
We propose two algorithms that construct the new \textcolor{black}{features} as conjunctions of the initial primitive \textcolor{black}{features} or their negations.
The generated feature sets have reduced correlations between features and succeed in catching some of the hidden relations between individuals in a dataset.
For example, a feature like $sky \wedge \neg building \wedge panorama$ would be true for non-urban images and is more informative than simple features expressing the presence or the absence of an object.
The notion of Pareto optimality is used to evaluate feature sets and to obtain a balance between total correlation and the complexity of the resulted feature set.
Statistical hypothesis testing is used in order to automatically determine the values of the parameters used for constructing a data-dependent feature set.
We experimentally show that our approaches achieve the construction of informative feature sets for multiple datasets.
\end{abstract}

\section{Introduction}

Most machine learning algorithms use a representation space based on a \textit{\textcolor{black}{feature}-based} format.
This format is a simple way to describe an instance as a measurement vector on a set of predefined \textcolor{black}{features}.
In the case of supervised learning, a class label is also available.
One limitation of the \textcolor{black}{feature}-based format is that supplied features sometimes do not adequately describe, in terms of classification, the semantics of the dataset.
This happens, for example, when general-purpose \textcolor{black}{features} are used to describe a collection that contains certain relations between individuals.

In order to obtain good results in classification tasks, many algorithms and preprocessing techniques (e.g., SVM~\citep{COR95}, PCA~\citep{DUN89} \textit{etc.}) deal with non-adequate variables by internally changing the description space.
The main drawback of these approaches is that they function as a black box, where the new representation space is either hidden (for SVM) or completely synthetic and incomprehensible to human readers (PCA).

The purpose of our work is to construct a new feature set that is more descriptive for both supervised and unsupervised classification tasks.
In the same way that \textbf{frequent itemsets} \citep{PIA91} help users to understand the patterns in transactions, our goal with the new \textcolor{black}{features} is to help understand relations between individuals of datasets.
Therefore, the new features should be easily comprehensible by a human reader.
Literature proposes algorithms that construct features based on the original user-supplied \textcolor{black}{features} (called primitives).
However, to our knowledge, all of these algorithms construct the feature set in a supervised way, based on the class information, supplied \textit{a priori} with the data.

In order to construct new features, we propose two algorithms that create new feature sets in the absence of classified examples, in an unsupervised manner.
The first algorithm is an adaptation of an established supervised algorithm, making it unsupervised.
For the second algorithm, we have developed a completely new heuristic that selects, at each iteration, pairs of highly correlated \textcolor{black}{features} and replaces them with conjunctions of literals that do not co-occur.
Therefore, the overall redundancy of the feature set is reduced.
Later iterations create more complex Boolean formulas, which can contain negations (meaning absence of features).
We use statistical considerations (hypothesis testing) to automatically determine the value of parameters depending on the dataset, and a \textit{Pareto front}~\citep{SAW85}-inspired method for the evaluation.
The main advantage of the proposed methods over PCA or the kernel of the SVM is that the newly-created features are comprehensible to human readers (features like $ people \wedge manifestation \wedge urban$ and $people \wedge \neg urban \wedge forest$ are easily interpretable).

In Sections~\ref{sec:ufringe}~and~\ref{sec:our-proposition}, we present our proposed algorithms and in Section~\ref{sec:OI} we describe the evaluation metrics and the complexity measures.
In Section~\ref{sec:xps}, we perform a set of initial experiments and outline some of the inconveniences of the algorithms.
In Section~\ref{sec:improvements}, by use of statistical hypothesis testing, we address these weak points, notably the choice of the threshold parameter.
In Section~\ref{sec:second-xp}, a second set of experiments validates the proposed improvements.
Finally, Section~\ref{sec:conclusion-future-work} draws the conclusion and outlines future works.

\subsection{Motivation: why construct a new feature set}
\label{sec:when-to-construct}

\begin{figure}[h]
  \centering
  \begin{tabular}{c@{}c@{}c}
  	\subfloat{\label{sub-fig:tags_gANDr1}\includegraphics[width=0.32\textwidth]{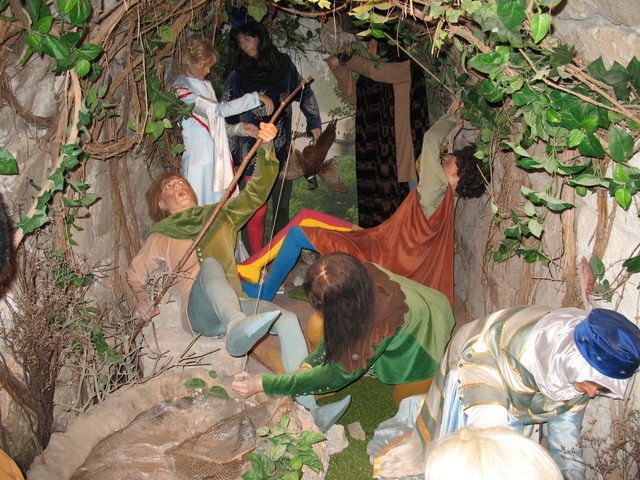}} & 
  	\subfloat{\label{sub-fig:tags_gANDr2}\includegraphics[width=0.32\textwidth]{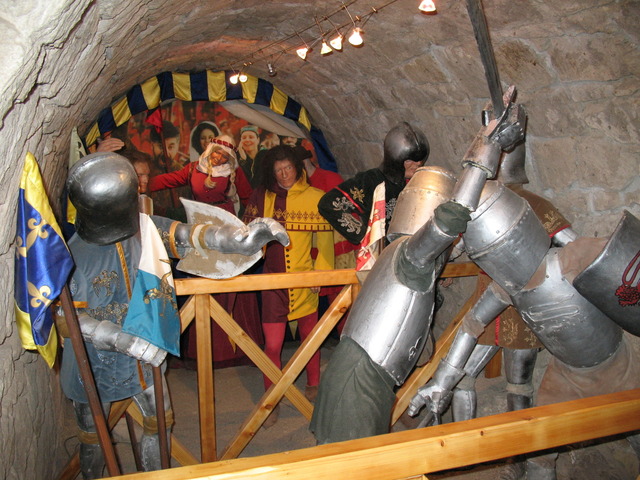}} & 
  	\subfloat{\label{sub-fig:tags_gANDr3}\includegraphics[width=0.32\textwidth]{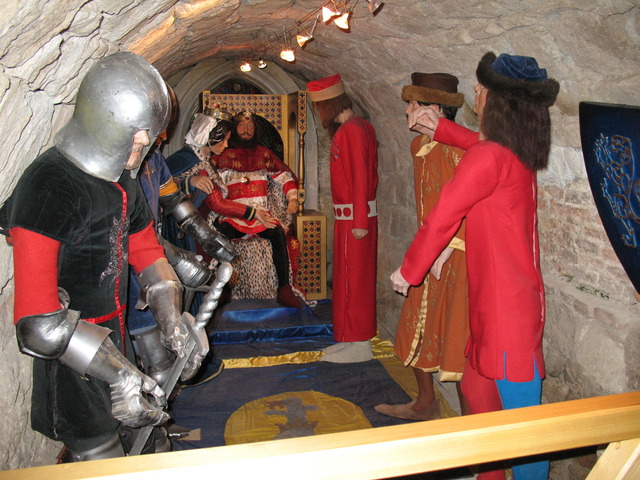}} \\[-0.5cm]
  	
  	\subfloat{\label{sub-fig:tags_gNOTr1}\includegraphics[width=0.32\textwidth]{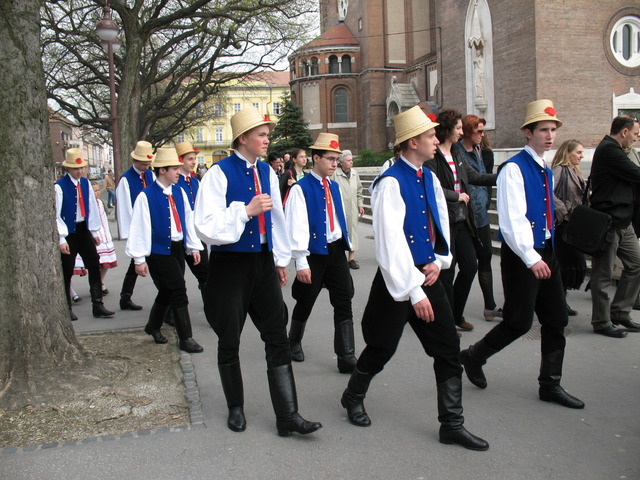}} &
  	\subfloat{\label{sub-fig:tags_gNOTr2}\includegraphics[width=0.32\textwidth]{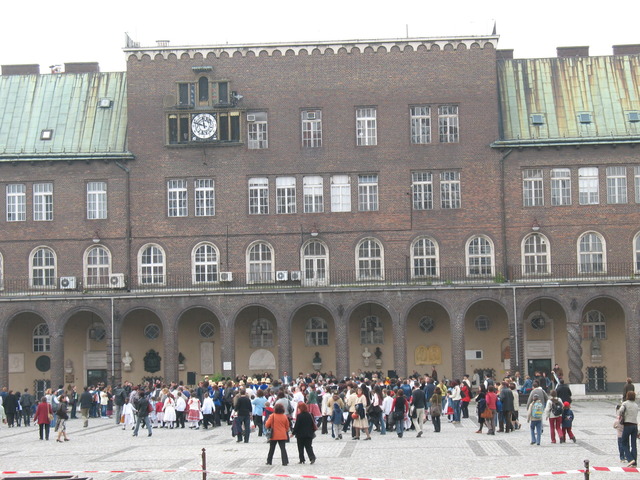}} &
  	\subfloat{\label{sub-fig:tags_gNOTr3}\includegraphics[width=0.32\textwidth]{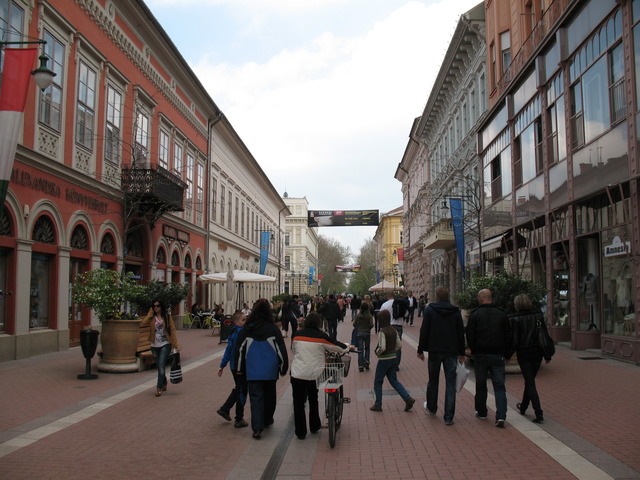}} \\[-0.4cm]
  \end{tabular}
  
	\caption{Example of images tagged with $\{ groups$, $road$, $building$, $interior\}$}
  	\label{fig:tags-examples}
\end{figure}

In the context of classification (supervised or unsupervised), a useful feature needs to portray new information.
A feature $p_j$, that is highly correlated with another feature $p_i$, does not bring any new information, since the value of $p_j$ can be deduced from that of $p_i$.
Subsequently, one could filter out ``irrelevant'' \textcolor{black}{features} before applying the classification algorithm.
But by simply removing certain features, one runs the risk of losing important information of the \textbf{hidden structure of the feature set}, and this is the reason why we perform \textbf{feature construction}.
\textcolor{black}{Feature construction attempts to increase the expressive power of the original features by discovering missing information about relationships between features.}

We deal primarily with datasets described with \textbf{Boolean} \textcolor{black}{features}.
Any dataset described by using the \textcolor{black}{feature}-value format can be converted to a binary format using discretization and binarization.
In real-life datasets, most binary \textcolor{black}{features} have specific meanings.
Let us consider the example of a set of images that are tagged using Boolean features.
Each feature marks the presence (\textbf{true}) or the absence (\textbf{false}) of a certain object in the image.
These objects could include: $water$, $cascade$, $manifestation$, $urban$, $groups$ or $interior$.
In this case, part of the semantic structure of the feature set can be guessed quite easily.
Relations like ``is-a'' and ``part-of'' are fairly intuitive: $cascade$ is a sort of $water$, $paw$ is part of $animal$ etc.
But other relations might be induced by the semantics of the dataset (images in our example).
$manifestation$ will co-occur with $urban$, for they usually take place in the city.
Fig.~\ref{fig:tags-examples} depicts a simple image dataset described using the feature set $\{ groups$, $road$, $building$, $interior\}$.
The feature set is quite redundant and some of the features are non-informative (e.g., feature $groups$ is present for all individuals).
Considering co-occurrences between features, we could create the more eloquent features $groups \wedge \neg road \wedge interior$ (describing the top row) and $groups \wedge road \wedge building$ (describing the bottom row).

The idea is to create a data-dependent feature set, so that the new features are as independent as possible, limiting co-occurrences between the new features.
At the same time they should be comprehensible to the human reader.

\subsection{Related work}
\label{sec:state-of-the-art}

\textcolor{black}{The literature proposes methods for augmenting the descriptive power of features.
\citet{LIU98} collects some of them and divides them into three categories: feature selection, feature extraction and feature construction.
}

\textcolor{black}{\textbf{Feature selection}~\citep{LAL00,MO11} seeks to filter the original feature set in order to remove redundant features.
This results in a representation space of lower dimensionality.
\textbf{Feature extraction} is a process that extracts a set of new features from the original features through functional mapping~\citep{MOT02}. }
\textcolor{black}{For example, t}he \textbf{SVM algorithm}~\citep{COR95} constructs a kernel function that changes the description space into a new separable one.
Supervised and non-supervised algorithms can be boosted by pre-processing with \textbf{principal component analysis} (PCA)~\citep{DUN89}.
PCA is a mathematical procedure that uses an orthogonal transformation to convert a set of observations of possibly correlated variables into a set of values of uncorrelated variables, called \textit{principal components}.
\textbf{Manifold learning}~\citep{HUO05} can be seen as a classification approach where the representation space is changed internally in order to boost the performances.
Feature extraction mainly seeks to reduce the description space and redundancy \textcolor{black}{between features}.
Newly-created features \textcolor{black}{are rarely comprehensible} and very difficult to interpret.
\textcolor{black}{Both feature selection and feature extraction are inadequate for detecting relations between the original features.}

\textbf{Feature Construction} \textcolor{black}{is a process that discovers missing information about the relationships between features and augments the space of features by inferring or creating additional features~\citep{MOT02}.
This usually results in a representation space with a larger dimension than the original space.
Constructive induction~\citep{MIC83} is a process of constructing new features using two intertwined searches~\citep{BLOE98}: one in the representation space (modifying the feature set) and another in the hypothesis space (using classical learning methods).
The actual feature construction is done using a set of constructing operators and the resulted features are often conjunctions of primitives, therefore easily comprehensible to a human reader.
}
Feature construction has mainly been used with decision tree learning.
New \textcolor{black}{features} served as hypotheses and were used as discriminators in decision trees. 
Supervised feature construction can also be applied in other domains, like decision rule learning~\citep{ZHE95}.

\begin{algorithm}
\caption{General feature construction schema.}          
\label{algo:general-algorithm}             

\begin{algorithmic}                    

\REQUIRE $P$ -- set of primitive user-given features
\REQUIRE $I$ -- the data expressed using $P$ which will be used to construct features \\
\textbf{Inner parameters:} $Op$ -- set of operators for constructing features, $M$ -- machine learning algorithm to be employed
\ENSURE $F$ -- set of new (constructed and/or primitives) features.

\STATE $ F \gets P $
\STATE $ iter \gets 0$
\REPEAT
	\STATE $ iter \gets iter + 1$
	\STATE $I_{iter} \gets $ \textbf{convert}($I_{iter-1}, F$)
	\STATE $ output \gets $ Run \textbf{M}($I_{iter},F$)
	\STATE $ F \gets F \bigcup $  new feat. constructed with \textbf{Op}$(F, output)$
	\STATE prune useless features in $F$
\UNTIL{ stopping criteria are met.}

\end{algorithmic}
\end{algorithm}

Algorithm~\ref{algo:general-algorithm}, presented in \citet{GOM02,YAN91}, represents the general schema followed by most constructive induction algorithms. 
The general idea is to start from $I$, the dataset described with the set of primitive \textcolor{black}{features}.
Using a set of constructors and the results of a machine learning algorithm, the algorithm constructs new features that are added to the feature set.
In the end, useless features are pruned.
These steps are iterated until some stopping criterion is met (e.g., a maximum number of iterations \textcolor{black}{performed or a maximum number of }created features).

Most constructive induction systems construct \textcolor{black}{features} as conjunctions or disjunctions of literals.
Literals are the features or their negations. 
E.g., for the feature set $\{a,b\}$ the literal set is $\{a,\neg a,b, \neg b\}$.
Operator sets $\{ AND, Negation\}$ and $\{ OR, Negation\}$ are both complete sets for the Boolean space. 
Any Boolean function can be created using only operators from one set.
FRINGE~\citep{PAG90} creates new \textcolor{black}{features} using a decision tree that it builds at each iteration.
New \textcolor{black}{features} are conjunctions of the last two nodes in each positive path (a positive path connects the root with a leaf having the class label \textbf{true}).
The newly-created features are added to the feature set and then used in the next iteration to construct the decision tree.
This first algorithm of feature construction was initially designed to solve replication problems in decision trees.

Other algorithms have further improved this approach.
CITRE~\citep{MAT90} adds other search strategies like \textit{root} (selects first two nodes in a positive path) or \textit{root-fringe} (selects the first and last node in the path).
It also introduces domain-knowledge by applying filters to prune the constructed features.
CAT~\citep{ZHE98} is another example of a hypothesis-driven constructive algorithm similar to FRINGE.
It also constructs conjunctive features based on the output of decision trees.
It uses a dynamic-path based approach (the conditions used to generate new features are chosen dynamically) and it includes a pruning technique.

There are alternative representations, other than conjunctive and disjunctive.
The M-of-N and X-of-N representations use \textcolor{black}{feature}-value pairs.
An \textcolor{black}{feature}-value pair $AV_k (A_i=V_{ij})$ is \textbf{true} for an instance if and only if the \textcolor{black}{feature} $A_i$ has the value $V_{ij}$ for that instance.
The difference between M-of-N and X-of-N is that, while the second one counts the number of true \textcolor{black}{feature}-value pairs, the first one uses a threshold parameter to assign a value of truth for the entire representation.
The algorithm $ID2-of-3$~\citep{MUR91} uses M-of-N representations for the newly-created features.
It has a specialization and a generalization construction operator and it does not need to construct a new decision tree at each step, but instead integrates the feature construction in the decision tree construction.
The $XofN$~\citep{ZHE95} algorithm functions similarly, except that it uses the X-of-N representation.
It also takes into account the complexity of the features generated.

Comparative studies like \citet{ZHE96} show that conjunctive and disjunctive representations have very similar performances in terms of prediction accuracy and theoretical complexity.
M-of-N, while more complex, has a stronger representation power than the two before.
The X-of-N representation has the strongest representation power, but the same studies show that it suffers from data fragmenting more than the other three.

The problem with all of these algorithms is that they all work in a supervised environment and they cannot function without a class label.
In the following sections, we will propose two approaches towards unsupervised feature construction.

\section{uFRINGE - adapting FRINGE for unsupervised learning}
\label{sec:ufringe}

We propose \textbf{uFRINGE}, an unsupervised version of FRINGE, one of the first feature construction algorithms.
FRINGE~\citep{PAG90} is a framework algorithm (see Section~\ref{sec:state-of-the-art}), following the same general schema shown in Algorithm~\ref{algo:general-algorithm}.
It creates new features using a logical decision tree, created using a traditional algorithm like ID3~\citep{QUI86} or C4.5~\citep{QUI93}.
Taking a closer look at FRINGE, one would observe that its only component that is supervised is the decision tree construction.
The actual construction of features is independent of the existence of a class attribute.
Hence, using an unsupervised decision tree construction algorithm renders FRINGE unsupervised.

\textbf{Clustering trees}~\citep{BLO98} were introduced as generalized logical decision trees.
They are constructed using a top-down strategy.
At each step, the cluster under a node is split into two, seeking to maximize the intra-cluster variance.
The authors argue that supervised indicators, used in traditional decision trees algorithms, are special cases of intra-cluster variance, as they measure intra-cluster \textbf{class} diversity.
Following this interpretation, clustering trees can be considered generalizations of decision trees and are suitable candidates for replacing ID3 in \textbf{uFRINGE}.

Adapting FRINGE to use clustering trees is straightforward: it is enough to replace \textbf{M} in Algorithm~\ref{algo:general-algorithm} with the clustering trees algorithm.
At each step, uFRINGE constructs a clustering tree using the dataset and the current feature set.
Just like in FRINGE, new \textcolor{black}{features} are created using the conditions under the last two nodes in each path connecting the root to a leaf.
FRINGE constructs new features starting only from positive leaves (leaves labelled true).
But unlike decision trees, in classification trees the leaves are not labelled using class \textcolor{black}{features}.
Therefore, uFRINGE constructs new features based on all paths from root to a leaf.

Newly-constructed \textcolor{black}{features} are added to the feature set and used in the next classification tree construction.
The algorithm stops when either no more features can be constructed from the clustering tree or when a maximum allowed number of features have already been constructed.
%

\textcolor{black}{  \textbf{Limitations.}
uFRINGE is capable of constructing new features in an unsupervised context.
It is also relatively simple to understand and implement, as it is based on the same framework as FRINGE.
However, it suffers from a couple of drawbacks.
Constructed features tend to be redundant and contain doubles.
Newly-constructed features are added to the feature set and are used, alongside old features, in later iterations. 
Older features are never removed from the feature set and they can be combined multiple times, thus resulting in doubles in the constructed feature set.
What is more, old features can be combined with new features in which they already participated, therefore constructing redundant features (e.g., $f_2$ and $f_1 \wedge f_2 \wedge f_3$ resulting in $f_2 \wedge f_1 \wedge f_2 \wedge f_3$).
Another limitation is controlling the number of constructed features.
The algorithm stops when a maximum number of features are constructed.
This is very inconvenient, as the dimension of the new feature set cannot be known in advance and is highly dependent on the dataset.
Furthermore, constructing too many features leads to overfitting and an overly complex feature set.
These shortcomings could be corrected by refining the constructing operator and by introducing a filter operator.
}

\section{uFC - a greedy heuristic}
\label{sec:our-proposition}

We address the limitations of uFRINGE by proposing a second, innovative approach.
We propose an iterative algorithm that reduces the overall correlation of features of a dataset by iteratively replacing pairs of highly correlated \textcolor{black}{features} with conjunctions of literals.
We use a greedy search strategy to identify the features that are highly correlated, then use a construction operator to create new features.
From two correlated features $f_i$ and $f_j$ we create three new features: $f_i \wedge f_j$, $f_i \wedge \overline{f_j}$ and $\overline{f_i} \wedge f_j$.
In the end, both $f_i$ and $f_j$ are removed from the feature set.
The algorithm stops \textcolor{black}{when no more new features are created or when} it has performed a maximum number of iterations.
The \textcolor{black}{formalization and the} different key parts of the algorithm (e.g., the search strategy,construction operators or feature pruning) will be presented in the next sections.

\begin{figure}[!t]
  \centering
  \subfloat[] {
  	\label{sub-fig:step1}
  	\includegraphics[width=0.32\textwidth]{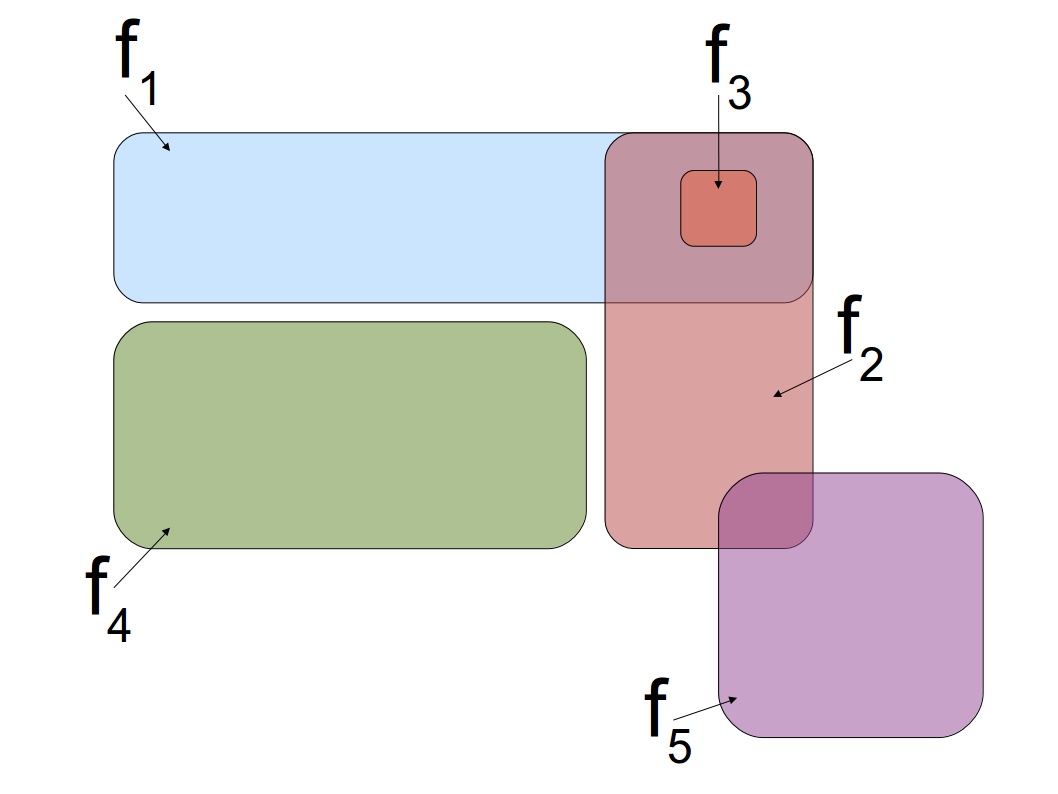}
  }         
  \hfill
  \subfloat[]{
  	\label{sub-fig:step2}
  	\includegraphics[width=0.32\textwidth]{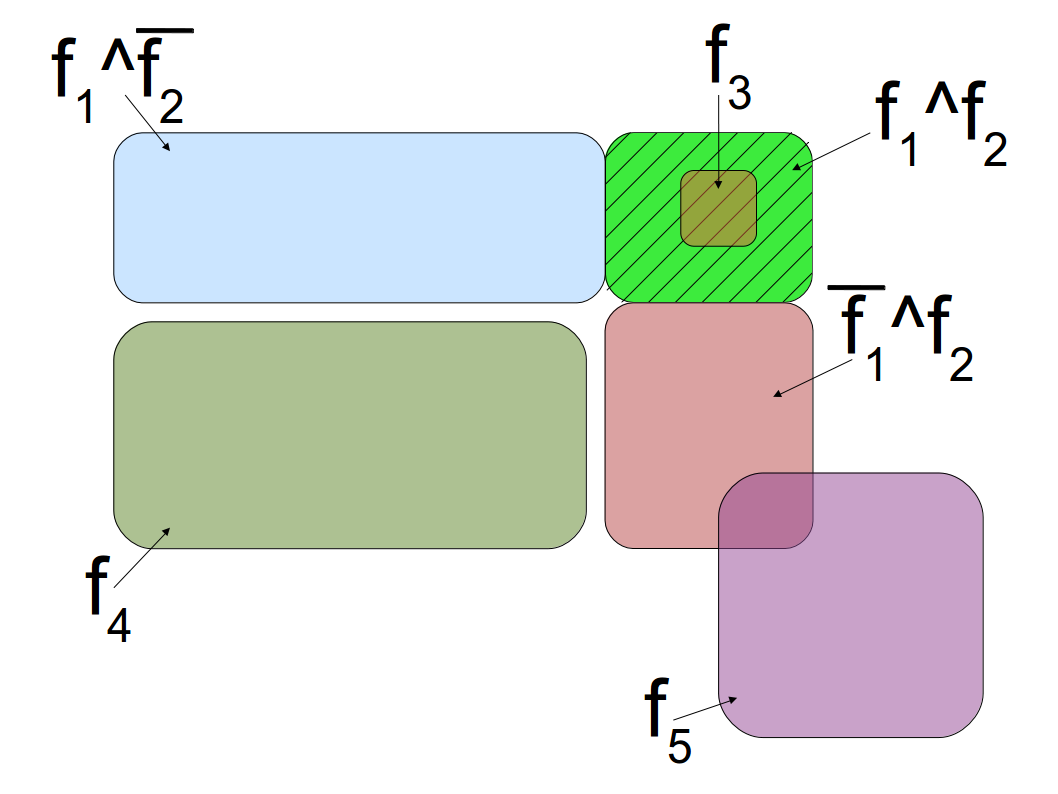}
  }
  \hfill
  \subfloat[]{
  	\label{sub-fig:step3}
  	\includegraphics[width=0.32\textwidth]{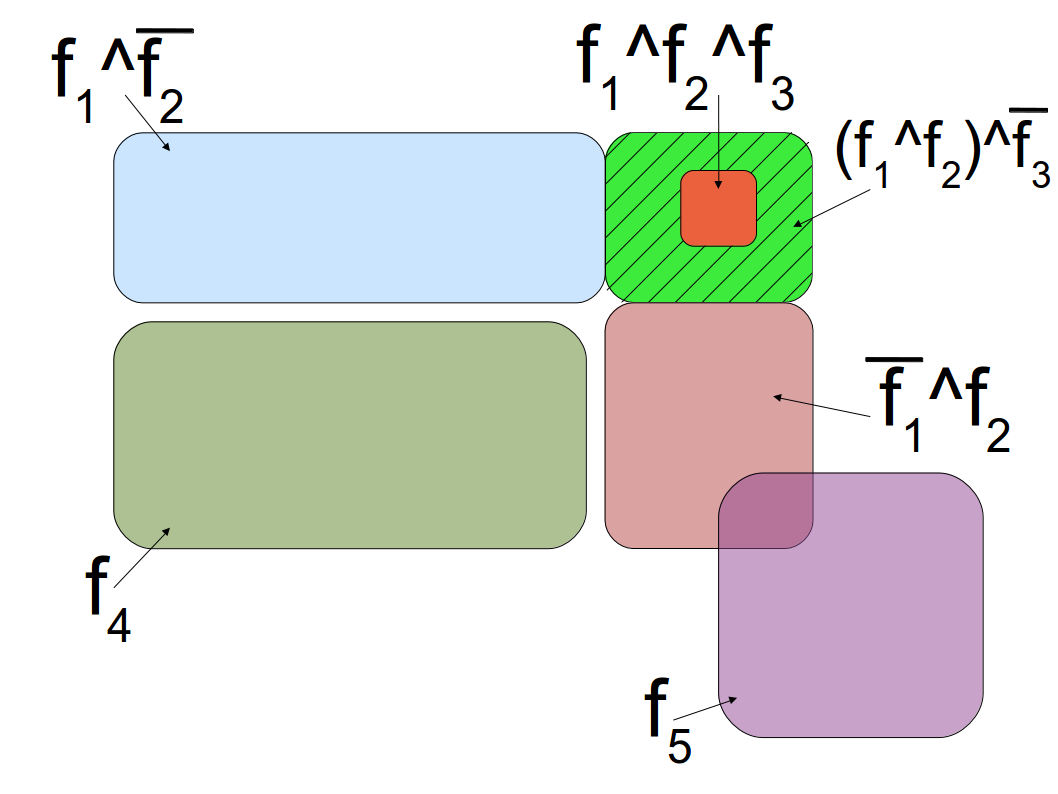}
  }
  
  \caption{Graphical representation of how new features are constructed - Venn diagrams. (a) Iter. 0: Initial \textcolor{black}{features} (Primitives), (b) Iter. 1: Combining $f_1$ and $f_2$ and (c) Iter. 2: Combining $f_1 \wedge f_2$ and $f_3$}
  \label{fig:algo-evolution}
\end{figure}

Fig.~\ref{fig:algo-evolution} illustrates visually, using Venn diagrams, how the algorithm replaces the old features with new ones.
Features are represented as rectangles, where the rectangle for each feature contains the individuals having that feature set to \textbf{true}.
Naturally, the individuals in the intersection of two rectangles have both features set to \textbf{true}.
$f_1$ and $f_2$ have a big intersection, showing that they co-occur frequently.
On the contrary, $f_2$ and $f_5$ have a small intersection, suggesting that their co-occurrence is less than that of the hazard (negatively correlated).
$f_3$ is included in the intersection of $f_1$ and $f_2$, while $f_4$ has no common elements with any other.
$f_4$ is incompatible with all of the others.

In the first iteration, $f_1$ and $f_2$ are combined and 3 features are created: $f_1 \wedge f_2$, $f_1 \wedge \overline{f_2}$ and $\overline{f_1} \wedge f_2$.
These new features will replace \textcolor{black}{$f_1$ and $f_2$,} the original ones.
At the second iteration, $f_1 \wedge f_2$ is combined with $f_3$.
As $f_3$ is contained in $f_1 \wedge f_2$, the \textcolor{black}{feature}  $\overline{f_1 \wedge f_2} \wedge f_3$ will have a support equal to zero and will be removed.
Note that $f_2$ and $f_5$ are never combined, as they are considered uncorrelated.
The final feature set will be $\{f_1 \wedge \overline{f_2}, f_1 \wedge f_2 \wedge f_3, f_1 \wedge f_2 \wedge \overline{f_3}, \overline{f_1} \wedge f_2, f_4, f_5 \} $

\subsection{\textbf{uFC} - the proposed algorithm}
\label{subsec:formalisation-algorithm}

We define the set $P = \{ p_1, p_2, ... , p_k \}$ of $k$ user-supplied \textcolor{black}{initial} features and $I = \{ i_1, i_2, ... , i_n \}$ the dataset described using $P$.
We start from the hypothesis that even if the primitive set $P$ cannot adequately describe the dataset $I$, there is a data-specific feature set $ F = \{ f_1, f_2, ... , f_m \} $ that can be created in order to represent the data better.
New features are iteratively created using conjunctions of primitive features or their negations (as seen in Fig.~\ref{fig:algo-evolution}).
Our algorithm does not use the output of a learning algorithm in order to create the new features.
Instead we use a \textcolor{black}{greedy search strategy} and a feature set evaluation function that can determine if a newly-obtained feature set is more appropriate than the former one.

\begin{algorithm}
\caption{\textbf{uFC} - Unsupervised feature construction}
\label{algo:our-proposition}

\begin{algorithmic}
\REQUIRE $P$ -- set of primitive user-given features
\REQUIRE $I$ -- the data expressed using $P$ which will be used to construct features
\STATE \textbf{Inner parameters:} $\lambda$ -- correlation threshold for searching, $limit\_iter$ -- max no of iterations.
\ENSURE $F$ -- set of newly-constructed features.

\STATE $ F_{0} \gets P $
\STATE $ iter \gets 0$
\REPEAT

	\STATE $ iter \gets iter + 1$
	\STATE $ O \gets $ \textbf{search\_correlated\_pairs}($I_{iter}, F_{iter-1}, \lambda$)
	\STATE $ F_{iter} \gets F_{iter-1}$
	\WHILE{ $O \neq \emptyset $ }
    	\STATE $ pair \gets $ \textbf{highest\_scoring\_pair}($O$)
    	\STATE $ F_{iter} \gets F_{iter} \bigcup $ \textbf{construct\_new\_feat}($pair$)
    	\STATE \textbf{remove\_candidate}($O, pair$)
	\ENDWHILE	
	
	\STATE \textbf{prune\_obsolete\_features}($F_{iter}, I_{iter}$)
	\STATE $I_{iter+1} \gets $ \textbf{convert}($I_{iter}, F_{iter}$)
		
\UNTIL{ $ F_{iter} = F_{iter-1}$ \textbf{OR} $iter = limit\_iter$}

\STATE $ F \gets F_{iter}$

\end{algorithmic}
\end{algorithm}

The schema of our proposal is presented in Algorithm~\ref{algo:our-proposition}.
The feature construction is performed starting from the dataset $I$ and the primitives $P$.
The algorithm follows the general inductive schema presented in Algorithm~\ref{algo:general-algorithm}.
At each iteration, \textbf{uFC} searches for frequently co-occurring pairs in the feature set created at the previous iteration ($F_{iter-1}$).
It determines the candidate set $O$ and then creates new features as conjunctions of the highest scoring pairs.
The new features are added to the current set ($F_{iter}$), after which the set is filtered in order to remove obsolete features.
At the end of each iteration, the dataset $I$ is translated to reflect the feature set $F_{iter}$.
A new iteration is performed as long as new features were generated in the current iteration and a maximum number of iterations have not yet been reached ($limit\_iter$ is a parameter for the algorithm).

\subsection{Searching co-occurring pairs}
\label{subsec:new-feat-search}

The \textbf{search\_correlated\_pairs} function searches for frequently co-occurring pairs of features in a feature set $F$.
We start with an empty set $ O \gets \emptyset $ and we investigate all possible pairs of features $ \{f_i,f_j\} \in F \times F $.
We use a function ($r$) to measure the co-occurrence of a pair of features $\{f_i,f_j\}$ and compare it to a threshold $\lambda$.
If the value of the function is above the threshold, then their co-occurrence is considered as significant and the pair is added to $O$.
Therefore, $O$ will be
$$ O = \{ \{f_i,f_j\} | \forall \{f_i,f_j\} \in F \times F \, so \, that \, \textbf{r}(\{f_i,f_j\}) > \lambda \} $$

The $r$ function is the empirical \textbf{Pearson correlation coefficient}, which is a measure of the strength of the linear dependency between two variables.
$r \in [-1, 1]$ and it is defined as the covariance of the two variables divided by the product of their standard deviations.
The sign of the $r$ function gives the direction of the correlation (inverse correlation for $r<0$ and positive correlation for $r>0$), while the absolute value or the square gives the strength of the correlation.
A value of 0 implies that there is no linear correlation between the variables.
When applied to Boolean variables, having the contingency table as shown in Table~\ref{tab:contingency-Table}, the $r$ function has the following formulation:
$$ r(\{f_i,f_j\}) = \frac{a \times d - b \times c}{\sqrt{(a+b) \times (a+c) \times (b+d) \times (c+d)}} $$

\begin{table}[htbp]
\caption{Contingency table for two Boolean \textcolor{black}{features}}
\label{tab:contingency-Table}

\centering
\begin{tabular}{lll}
\toprule
 & $f_j$ & $\neg f_j$ \\ \midrule
$f_i$ & a & b \\ 
$\neg f_i$ & c & d \\ \bottomrule
\end{tabular}

\end{table}

The $\lambda$ threshold parameter will serve to fine-tune the number of selected pairs.
Its impact on the behaviour of the algorithm will be studied in Section~\ref{subsec:pareto-evaluation}.
A method of automatic choice of $\lambda$ using statistical hypothesis testing is presented in Section~\ref{subsec:choose-lambda}.

\subsection{Constructing and pruning features}
\label{subsec:prunning-feature-set}

Once $O$ is constructed, \textbf{uFC} performs a greedy search.
The function \textbf{highest\_scoring\_pair} is iteratively used to extract from $O$ the pair $\{f_i,f_j\}$ that has the highest co-occurrence score.

The function \textbf{construct\_new\_feat} constructs three new features: $f_i \wedge f_j$, $\overline{f_i} \wedge f_j$ and $f_i \wedge \overline{f_j}$.
They represent, respectively, the intersection of the initial two features and the relative complements of one \textcolor{black}{feature} in the other.
The new features are guaranteed by construction to be negatively correlated.
If one of them is set to true for an individual, the other two will surely be false.
At each iteration, very simple features are constructed: conjunctions of two literals.
The creation of more complex and semantically rich features appears through the iterative process.
$f_i$ and $f_j$ can be either primitives or features constructed in previous iterations.

After the construction of features, the \textbf{remove\_candidate} function removes from $O$ the pair  $\{f_i,f_j\}$, as well as any other pair that contains $f_i$ of $f_j$.
When there are no more pairs in $O$, \textbf{prune\_obsolete\_features} is used to remove from the feature set two types of features:
\begin{itemize}
	\item \textbf{features that are false for all individuals}. 
	These usually appear in the case of hierarchical relations.
	\textcolor{black}{We consider that $f_1$ and $f_2$ have a hierarchical relation if all individuals that have feature $f_1$ true, automatically have feature $f_2$ true (e.g., $f_1$ ``is a type of'' $f_2$ or $f_1$ ``is a part of'' $f_2$).}
	One of the generated features \textcolor{black}{(in the example $\overline{f_1} \wedge f_2$)} is false for all individuals and, therefore, eliminated.
	In the example of $water$ and $cascade$, we create only $water \wedge cascade$ and $water \wedge \neg cascade$, since there cannot exist a cascade without water (considering that a value of \textbf{false} means the absence of a feature and not missing data).

	\item \textbf{features that participated in the creation of a new feature}. Effectively, all $\{ f_i | \{f_i, f_j\} \in O, f_j \in F\}$ are replaced by the newly-constructed features.
\end{itemize}%
$$ \{f_i, f_j |f_i, f_j \in F, \{f_i, f_j\} \in O \} \xrightarrow{repl.} \{ f_i \wedge f_j, \overline{f_i} \wedge f_j, f_i \wedge \overline{f_j}\}$$	

\section{Evaluation of a feature set}
\label{sec:OI}

To our knowledge, there are no widely accepted measures to evaluate the overall correlation between the \textcolor{black}{features} of a feature set.
We propose a measure inspired from the ``inclusion-exclusion'' principle~\citep{FEL50}.
In set theory, this principle permits to express the cardinality of the finite reunion of finite ensembles by considering the cardinality of those ensembles and their intersections.
In the Boolean form, it is used to estimate the probability of a \textit{clause} (disjunction of literals) as a function of its composing \textit{terms} (conjunctions of literals).

Given the feature set $F = \{f_1, f_2, ... , f_m \}$, we have:
$$
p(f_1 \vee f_2 \vee ... \vee f_m) = \sum_{k=1}^{m} \left( (-1)^{k-1} \sum_{1 \leq i_1<...<i_k \leq m} p(f_{i_1} \wedge f_{i_2} \wedge ... \wedge f_{i_k}) \right)
$$
which, by putting apart the first term, is equivalent to:
$$
p(f_1 \vee f_2 \vee ... \vee f_m) = \sum_{i=1}^{m} p(f_i) + \sum_{k=2}^{m} \left( (-1)^{k-1} \sum_{1 \leq i_1<...<i_k \leq m} p(f_{i_1} \wedge f_{i_2} \wedge ... \wedge f_{i_k}) \right) 
$$

Without loss of generality, we can consider that each individual has at least one feature set to \textbf{true}.
Otherwise, we can create an artificial feature ``null'' that is set to \textbf{true} when all the others are \textbf{false}.
Consequently, the left side of the equation is equal to 1.
On the right side, the second term is the probability of intersections of the features.
Knowing that $ 1 \leq \sum_{i=1}^{m} p(f_i) \leq m $, this probability of intersection has a value of zero when all features are incompatible (no overlapping).
It has a ``worst case scenario'' value of $m-1$, when all individuals have all the features set to \textbf{true}.

Based on these observations, we propose the \textbf{Overlapping Index} evaluation measure:
$$ OI(F) = \frac{\sum_{i=1}^{m} p(f_i) - 1}{m-1} $$
where $ OI(F) \in [0,1] $ and ``better'' towards zero.
Hence, a feature set $F_1$ describes a dataset better than another feature set $F_2$ when $ OI(F_1) < OI(F_2)$.

\subsection{Complexity of the feature set.} 
\label{subsec:complexity-tradeoff}

\textbf{Number of features} 
Considering the case of the majority of machine learning datasets, where the number of primitives is inferior to the number of individuals in the dataset, reducing correlations between features comes at the expense of increasing the number of features.
Consider the pair of features $\{ f_i, f_j\}$ judged correlated.
Unless $f_i \supseteq f_j$ or $f_i \subseteq f_j$, the algorithm will replace $\{ f_i, f_j\}$ by $\{ f_i \wedge f_j, \overline{f_i} \wedge f_j, f_i \wedge \overline{f_j}\}$, thus increasing the total number of features.
A feature set that contains too many \textcolor{black}{features} is no longer informative, nor comprehensible.
The maximum number of features that can be constructed is mechanically limited by the number of unique combinations of primitives in the dataset (the number of unique individuals).
$$ |F| \leq unique(I) \leq |I| $$
where $F$ is the constructed feature set and $I$ is the dataset.

To measure the complexity in terms of number of features, we use:
$$ C_0(F) = \frac{|F| - |P|}{unique(I) - |P|} $$
where $P$ is the primitive feature set.
$C_0$ measures the ratio between how many extra features are constructed and the maximum number of features that can be constructed.
$ 0 \leq C_0 \leq 1 $ and a value closer to $0$ means a feature set less complex.

\textbf{The average length of features} 
At each iteration, simple conjunctions of two literals are constructed.
Complex Boolean formulas are created by combining features constructed in previous iterations.
Long and complicated expressions generate incomprehensible features, which are more likely a random side-effect rather than a product of underlying semantics.

We define $C_1$ as the average number of literals (a primitive or its negation) that appear in a Boolean formula representing a new feature.
$$ \overline{P} = \{ \overline{p_i} | p_i \in P \}; \; \mathcal{L} = P \bigcup \overline{P} $$
$$ C_1(F) = \frac{\sum_{f_i \in F} |\{l_j | l_j \in \mathcal{L}, l_j \, appears \, in \, f_i \}|}{|F|} $$
where $P$ is the primitive set and $ 1 \leq C_1 < \infty $.

As more iterations are performed, the feature set contains more features ($C_0$ grows) which are increasingly more complex ($C_1$ grows).
This suggests a correlation between the two.
What is more, since $C_1$ can potentially double at each iteration and $C_0$ can have at most a linear increase, the correlation is exponential.
For this reason, in the following sections we shall use only $C_0$ as the complexity measure.

\textcolor{black}{\textbf{Overfitting}
All algorithms that learn from data risk overfitting the solution to the learning set.
There are two ways in which \textbf{uFC} can overfit the resulted feature set, corresponding to the two complexity measures above: a) constructing too many features (measure $C_0$) and b) constructing features that are too long (measure $C_1$).
The worst overfitting of type a) is when the algorithm constructs as many features as the maximum theoretical number (one for each individual in the dataset).
The worst overfitting of type b) appears in the same conditions, where each constructed feature is a conjunction of all the primitives appearing for the corresponding individual.
The two complexity measures can be used to quantify the two types of ovefitting.
Since $C_0$ and $C_1$ are correlated, both types of overfitting appear simultaneously and can be considered as two sides of a single phenomenon.
}

\subsection{The trade-off between two opposing criteria}
\label{subsec:pareto-front}

\textcolor{black}{$C_0$ is a measure of how overfitted a feature set is.
In order to avoid overfitting, feature set complexity should be kept at low values, while the algorithm optimizes the co-occurrence score of the feature set.}
Optimizing both the correlation score and the complexity at the same time is not possible\textcolor{black}{, as they are opposing criteria}.
A compromise between the two must be achieved.
This is equivalent to the optimization of two contrary criteria, which is a very well-known problem in multi-objective optimization.
To acquire a trade-off between the two mutually contradicting objectives, we use the concept of \textbf{Pareto optimality}~\citep{SAW85}, originally developed in economics.
Given multiple feature sets, a set is considered to be Pareto optimal if there is no other set that has both a better correlation score and a better complexity for a given dataset.
Pareto optimal feature sets will form the Pareto front.
This means that no single optimum can be constructed, but rather a class of optima, depending on the ratio between the two criteria.

\textcolor{black}{We plot the solutions in the plane defined by the complexity, as one axis, and the co-occurrence score, as the other.}
Constructing the Pareto front \textcolor{black}{in this plane} makes a visual evaluation of \textcolor{black}{several characteristics of the \textbf{uFC} algorithm possible,} based on the deviation of solutions compared to the front.
\textcolor{black}{The distance between the different solutions and the constructed Pareto front visually shows how stable the algorithm is.
The convergence of the algorithm can be visually evaluated by how fast (in number of performed iterations) the algorithm transits the plane from the region of solutions with low complexity and high co-occurrence score to solutions with high complexity and low co-occurrence.
We can visually evaluate overfitting, which corresponds to the region of the plane with high complexity and low co-occurence score.
Solutions found in this region are overfitted.}

\textcolor{black}{In order to avoid overfitting,} we propose the \textbf{``closest-point''} heuristic for finding a compromise between $OI$ and $C_0$.
We consider the two criteria to have equal importance.
\textcolor{black}{We consider as a good compromise, the solution in which the gain in co-occurrence score and the loss in complexity are fairly equal.}
If one of the indicators has a value considerably larger than the other, the solution is considered to be unsuitable.
Such solutions would have either a high correlation between features or a high complexity.
Therefore, we perform a battery of tests and we search \textit{a posteriori} the Pareto front for solutions for which the two indicators have essentially equal values.
In the space of solutions, this translates into a minimal Euclidian distance between the solution and the ideal point (the point $(0; 0)$).

\section{Initial Experiments}
\label{sec:xps}

\begin{figure}[!t]
  \centering
  \subfloat[]{
  	\label{sub-fig:tag1_ex1}
  	\includegraphics[width=0.32\textwidth]{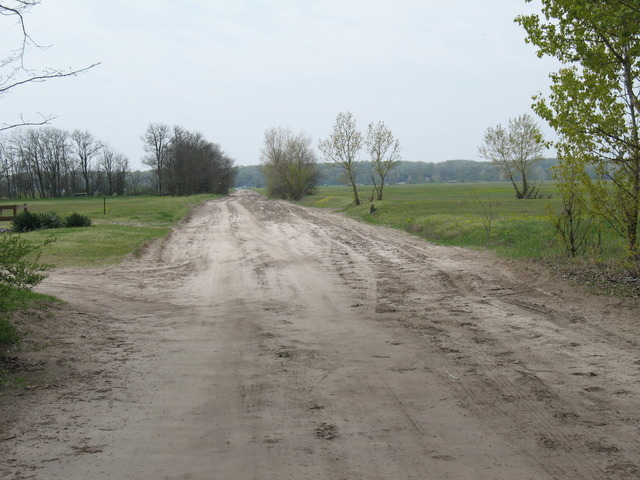}
  }   
  \hfill            
  \subfloat[]{
  	\label{sub-fig:tag1_ex2}
  	\includegraphics[width=0.32\textwidth]{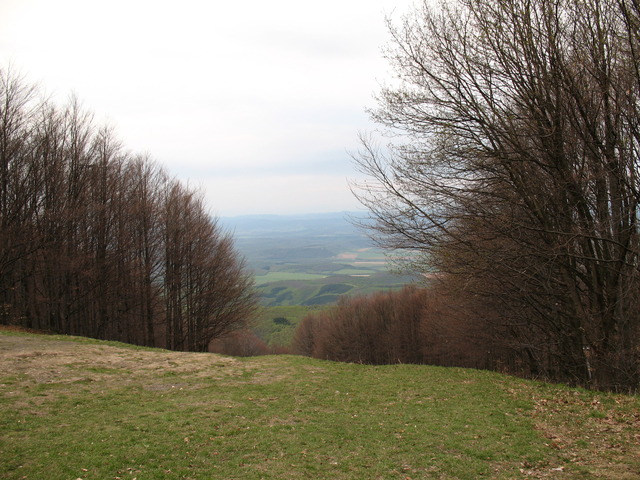}
  }
  \hfill
  \subfloat[]{
  	\label{sub-fig:tag1_ex3}
  	\includegraphics[width=0.32\textwidth]{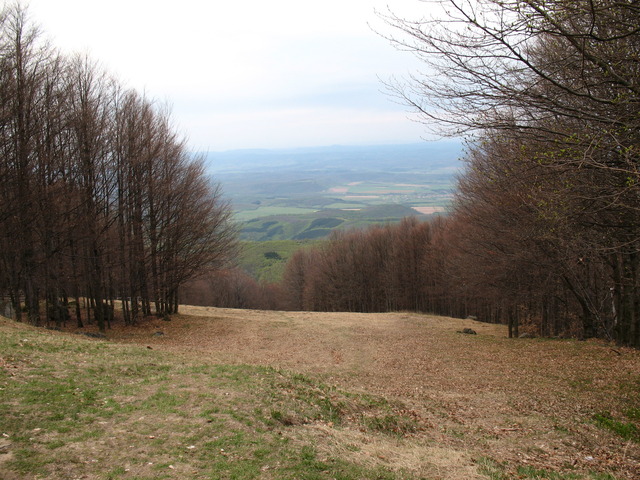}
  }
  
  \caption{Images related to the newly-constructed feature $sky \wedge \overline{building} \wedge panorama$ on \texttt{hungarian}: (a) Hungarian puszta, (b)(c) Hungarian M\'atra mountains}
  \label{fig:tag1-examples}
\end{figure}


Throughout the experiments, \textbf{uFC} was executed by varying only the two parameters: $\lambda$ and $limit_{iter}$.
We denote an execution with specific values for parameters as \textbf{uFC}$(\lambda$,~$limit_{iter})$, whereas the execution where the parameters were determined \textit{a posteriori} using the ``closest-point'' strategy will be noted \textbf{uFC*}$(\lambda$,~$limit_{iter})$.
For \textbf{uFRINGE}, the maximum number of features was set at 300.
We perform a comparative evaluation of the two algorithms seen from a qualitative and quantitative point of view, together with examples of typical executions.
Finally, we study the impact of the two parameters of \textbf{uFC}.

Experiments were performed on three Boolean datasets.
The \texttt{hungarian} dataset\footnote{\url{http://eric.univ-lyon2.fr/~arizoiu/files/hungarian.txt}} is a real-life collection of images, depicting Hungarian urban and countryside settings.
Images were manually tagged using one or more of the 13 tags.
Each tag represents an object that appears in the image (eg. tree, cascade etc.).
The tags serve as features and a feature takes the value \textbf{true} if the corresponding object is present in the image or \textbf{false} otherwise.
The resulted dataset contains 264 individuals, described by 13 Boolean features.
Once the dataset was constructed, the images were not used any more.
The \texttt{street} dataset\footnote{\url{http://eric.univ-lyon2.fr/~arizoiu/files/street.txt}} was constructed in a similar way, starting from images taken from the LabelMe dataset~\citep{RUS08}.
608 urban images from Barcelona, Madrid and Boston were selected.
Image labels were transformed into tags depicting objects by using the uniformization list provided with the toolbox.
The dataset contains 608 individuals, described by 66 Boolean features.

The third dataset is ``Spect Heart''\footnote{\url{http://archive.ics.uci.edu/ml/datasets/SPECT+Heart}} from the UCI.
The corpus is provided with a ``class'' attribute and divided into a learning corpus and a testing one.
We eliminated the class attribute and concatenated the learning and testing corpus into a single dataset.
It contains 267 instances described by 22 Boolean \textcolor{black}{features}.
Unlike the first two datasets, the features of \texttt{spect} have no specific meaning, being called ``F1'', ``F2'', ... , ``F22''.

\subsection{\textbf{uFC} and \textbf{uFRINGE}: Qualitative evaluation}
\label{subsec:qualitative-evaluation}

\begin{figure}[!t]
  \centering
  \subfloat[]{
  	\label{sub-fig:tag2_ex1}
  	\includegraphics[width=0.3747\textwidth]{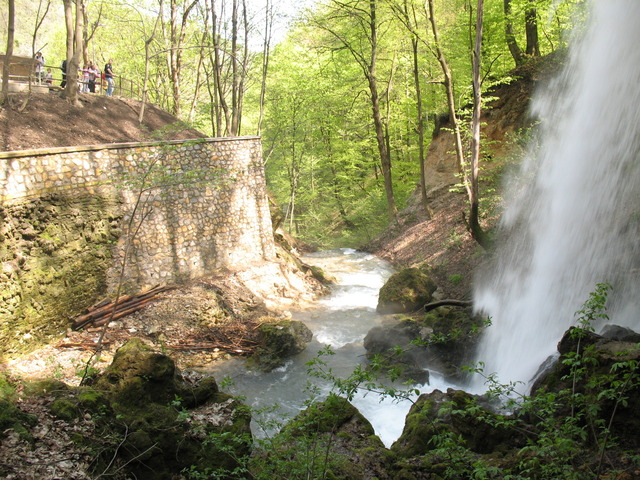}
  }    
  \hfill
  \subfloat[]{
  	\label{sub-fig:tag2_ex2}
  	\includegraphics[width=0.3747\textwidth]{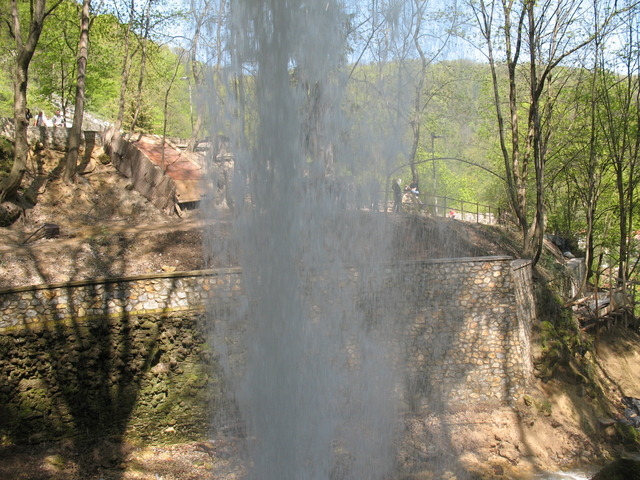}
  }
  \hfill
  \subfloat[]{
  	\label{sub-fig:tag2_ex3}
  	\includegraphics[width=0.2104\textwidth]{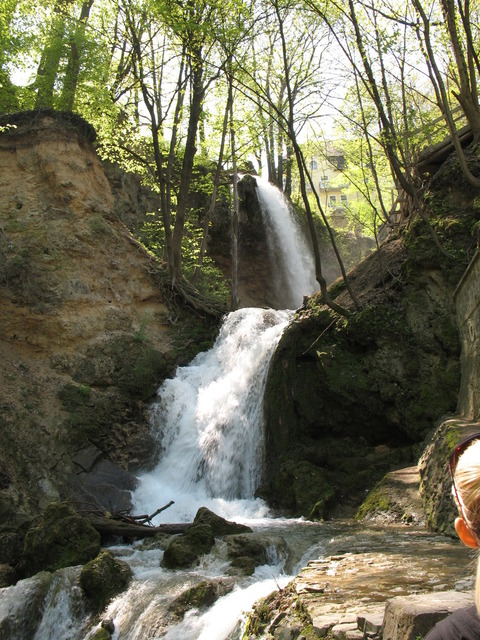}
  }
  
  \caption{Images related to the newly-constructed feature $water \wedge cascade \wedge tree \wedge forest$ on \texttt{hungarian}}
  \label{fig:tag2-examples}
\end{figure}

For the human reader, it is quite obvious why $water$ and $cascade$ have the tendency to appear together or why $road$ and $interior$ have the tendency to appear separately.
One would expect, that based on a given dataset, the algorithms would succeed in making these associations and catching the underlying semantics.
Table~\ref{tab:exec-results} shows the features constructed with \textbf{uFRINGE} and \textbf{uFC*}$(0.194,2)$ on \texttt{hungarian}.
A quick overview shows that constructed features manage to make associations that seem ``logical'' to a human reader.
For example, one would expect the feature $sky \wedge \overline{building} \wedge panorama$ to denote images where there is a panoramic view and the sky, but no buildings, therefore suggesting images outside the city.
Fig.~\ref{fig:tag1-examples} supports this expectation.
Similarly, the feature $\overline{sky} \wedge building \wedge groups \wedge road$ covers urban images, where groups of people are present and $water \wedge cascade \wedge tree \wedge forest$ denotes a cascade in the forest (Fig.~\ref{fig:tag2-examples}).

\begin{figure}[!t]
  \centering
  \subfloat[]{
  	\label{sub-fig:tag3_ex1}
  	\includegraphics[width=0.32\textwidth]{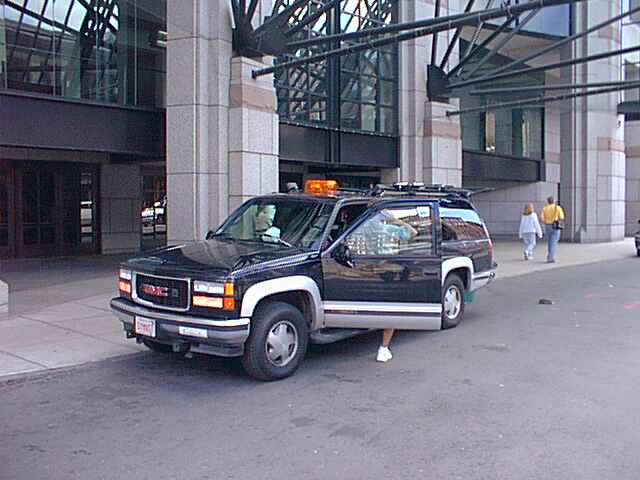}
  }       
  \hfill         
  \subfloat[]{
  	\label{sub-fig:tag3_ex2}
  	\includegraphics[width=0.32\textwidth]{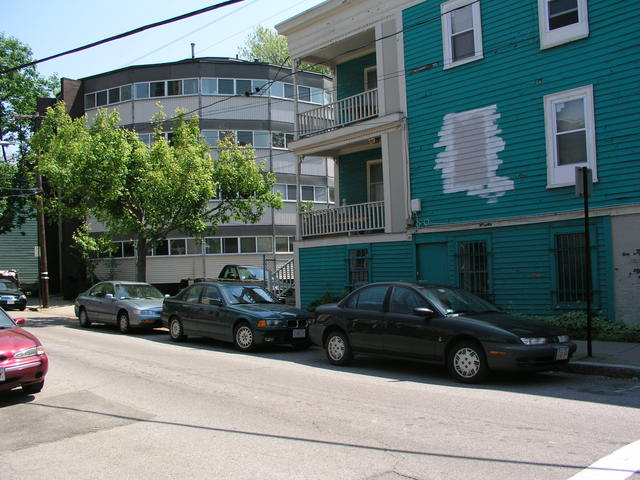}
  }
  \hfill
  \subfloat[]{
  	\label{sub-fig:tag3_ex3}
  	\includegraphics[width=0.32\textwidth]{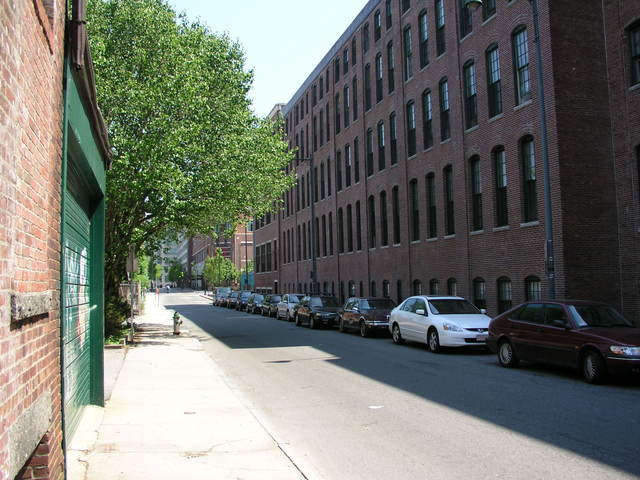}
  }
  
  \caption{Images related to the newly-constructed feature $headlight \wedge windshield \wedge \overline{arm} \wedge head$ on \texttt{street}}
  \label{fig:tag3-examples}
\end{figure}

Comprehension quickly deteriorates \textcolor{black}{when the constructed feature set is overfitted, }when the constructed features are too complex.
The execution of \textbf{uFC}$(0.184,5)$ reveals features like:
\begin{equation*}
\begin{split}
\overline{sky \wedge building} \wedge tree \wedge building \wedge\overline{forest} \wedge \overline{\overline{sky} \wedge building \wedge groups \wedge road} \wedge \\
\overline{sky \wedge \overline{building} \wedge \overline{panorama}}
\wedge \overline{\overline{\overline{groups} \wedge road} \wedge person}
\wedge \overline{sky} \wedge \overline{groups \wedge road} 
\end{split}
\end{equation*}
Even if the formula is not in the Disjunctive Normal Form (DNF), it is obvious that it is too complex to make any sense.
If \textbf{uFC} tends to construct overly complex features, \textbf{uFRINGE} suffers from another type of dimensionality curse.
Even if the complexity of features does not impede comprehension, the fact that there are over 300 hundred features constructed from 13 primitives makes the newly-constructed feature set unusable.
The number of features is actually greater than the number of individuals in the dataset, which proves that some of the features are redundant.
The actual \textcolor{black}{correlation score of} the newly-created feature set is even greater than the initial primitive set.
\textcolor{black}{What is more, new features present redundancy, just as predicted in section \ref{sec:ufringe}. 
For example, the feature $water \wedge forest \wedge grass \wedge water \wedge person$ which contains two times the primitive $water$.  }

The same conclusions are drawn from execution on the \texttt{street} dataset.
\textbf{uFC*}$(0.322, 2)$ creates comprehensible features.
For example $headlight \wedge windshield \wedge \overline{arm} \wedge head$ (Fig.~\ref{fig:tag3-examples}) suggests images in which the front part of cars appear.
It is especially interesting how the algorithm specifies $\overline{arm}$ in conjunction with $head$ in order to differenciate between people ($head \wedge arm$) and objects that have heads (but no arms).

\begin{table*}[tbp]
\caption{Feature sets constructed by \textbf{uFC} and \textbf{uFRINGE}}
\label{tab:exec-results}

\centering
\begin{tabular}{lll}
\toprule
\multicolumn{1}{c}{\textbf{primitives}} & \multicolumn{1}{c}{\textbf{uFRINGE}} & \multicolumn{1}{c}{\textbf{uFC(0.194, 2)}} \\ \midrule
$person$ & $ water \wedge forest \wedge grass \wedge water \wedge person$ & $groups \wedge \overline{road} \wedge interior$ \\ 
$groups$ & $ \overline{panorama} \wedge \overline{building} \wedge forest \wedge grass$ & $groups \wedge \overline{road} \wedge \overline{interior}$ \\ 
$water$ & $tree \wedge person \wedge grass$ & $ \overline{groups \wedge \overline{road}} \wedge interior$ \\ 
$cascade$ & $tree \wedge person \wedge \overline{grass}$ & $ water \wedge cascade \wedge tree \wedge forest$ \\ 
$ sky$ & $ \overline{groups} \wedge tree \wedge person$ & $ water \wedge cascade \wedge \overline{tree \wedge forest}$ \\ 
$ tree$ & $ \overline{person} \wedge interior$ & $ \overline{water \wedge cascade} \wedge tree \wedge forest$ \\ 
$ grass$ & $ \overline{person} \wedge \overline{interior}$ & $sky \wedge building \wedge tree \wedge \overline{forest}$ \\ 
$ forest$ & $ water \wedge \overline{panorama} \wedge grass \wedge \overline{groups} \wedge tree$ & $sky \wedge building \wedge \overline{tree \wedge \overline{forest}}$ \\ 
$ statue$ & $ water \wedge \overline{panorama} \wedge \overline{grass \wedge \overline{groups} \wedge tree}$ & $ \overline{sky \wedge building} \wedge tree \wedge \overline{forest}$ \\ 
$building$ & $ \overline{\overline{statue} \wedge \overline{groups}} \wedge groups$ & $sky \wedge \overline{building} \wedge panorama$ \\ 
$ road$ & $ \overline{\overline{statue} \wedge \overline{groups}} \wedge \overline{groups}$ & $sky \wedge \overline{building} \wedge \overline{panorama}$ \\ 
$ interior$ & $ \overline{panorama} \wedge \overline{statue} \wedge \overline{groups}$ & $ \overline{sky \wedge \overline{building}} \wedge panorama$ \\ 
$panorama$ & $ \overline{\overline{grass} \wedge \overline{water} \wedge \overline{forest}} \wedge sky$ & $ \overline{groups} \wedge road \wedge person$ \\ 
 & $ \overline{\overline{grass} \wedge \overline{water} \wedge \overline{forest}} \wedge \overline{sky}$ & $ \overline{groups} \wedge road \wedge \overline{person}$ \\ 
 & $person \wedge \overline{grass} \wedge \overline{water} \wedge \overline{forest}$ & $ \overline{\overline{groups} \wedge road} \wedge person$ \\ 
 & $groups \wedge \overline{sky} \wedge \overline{grass} \wedge \overline{building}$ & $ \overline{sky} \wedge building \wedge groups \wedge road$ \\ 
 & $groups \wedge \overline{sky} \wedge \overline{\overline{grass} \wedge \overline{building}}$ & $ \overline{sky} \wedge building \wedge \overline{groups \wedge road}$ \\ 
 & $ \overline{groups} \wedge \overline{person} \wedge \overline{water} \wedge \overline{forest} \wedge \overline{statue} \wedge \overline{groups}$ & $ \overline{\overline{sky} \wedge building} \wedge groups \wedge road$ \\ 
 & $ \overline{groups} \wedge \overline{\overline{person} \wedge \overline{water} \wedge \overline{forest} \wedge \overline{statue} \wedge \overline{groups}}$ & $ water \wedge \overline{cascade}$ \\ 
 & $ \overline{grass} \wedge \overline{person} \wedge \overline{statue}$ & $ \overline{tree} \wedge forest$ \\ 
 & $ \overline{grass} \wedge \overline{\overline{person} \wedge \overline{statue}}$ & $grass$ \\ 
 & \texttt{... and 284 others} & $statue$ \\
 \bottomrule
\end{tabular}
\end{table*}

\subsection{\textbf{uFC} and \textbf{uFRINGE}: Quantitative evaluation}
\label{subsec:quantitative-evaluation}

\begin{table}[htbp]
\caption{Values of indicators for multiple runs on each dataset}
\label{tab:result-all}

\centering
\begin{tabular}{llrrrr}
\toprule
 & \textbf{Strategy} & \textbf{$\#feat$} & \textbf{$length$} & \textbf{$OI$} & \textbf{$C_0$} \\ \midrule
 
\multirow{4}{*}{\begin{sideways}\texttt{hungar.}\end{sideways}} & \textbf{Primitives} & 13 & 1.00 & 0.24 & 0.00 \\ 
& \textbf{uFC*(0.194, 2)} & 21 & 2.95 & 0.08 & 0.07 \\ 
& \textbf{uFC(0.184, 5)} & 36 & 11.19 & 0.03 & 0.20 \\ 
& \textbf{uFRINGE} & 306 & 3.10 & 0.24 & 2.53 \\  \cmidrule{ 2- 6}

\multirow{4}{*}{\begin{sideways}\texttt{street}\end{sideways}} & \textbf{Primitives} & 66 & 1.00 & 0.12 & 0.00 \\ 
& \textbf{uFC*(0.446, 3)} & 81 & 2.14 & 0.06 & 0.04 \\ 
& \textbf{uFC(0.180, 5)} & 205 & 18.05 & 0.02 & 0.35 \\ 
& \textbf{uFRINGE} & 233 & 2.08 & 0.20 & 0.42 \\  \cmidrule{ 2- 6}

\multirow{4}{*}{\begin{sideways}\texttt{spect}\end{sideways}} & \textbf{Primitives} & 22 & 1.00 & 0.28 & 0.00 \\
& \textbf{uFC*(0.432, 3)} & 36 & 2.83 & 0.09 & 0.07 \\ 
& \textbf{uFC(0.218, 4)} & 62 & 8.81 & 0.03 & 0.20 \\ 
& \textbf{uFRINGE} & 307 & 2.90 & 0.25 & 1.45 \\ \bottomrule
\end{tabular}
\end{table}

Table~\ref{tab:result-all} shows, for the three datasets, the values of certain indicators, like the size of the feature set, the average length of a feature ($C_1$), the $OI$ and $C_0$ indicators.
For each dataset, we compare four feature sets: the initial feature set (primitives), the execution of \textbf{uFC*} (parameters determined by the ``closest-point'' heuristic), \textbf{uFC} with another random set of parameters and \textbf{uFRINGE}.
For the \texttt{hungarian} and \texttt{street} datasets, the same parameter combinations are used as in the qualitative evaluation. 

On all three datasets, \textbf{uFC*} creates feature sets that are less correlated than the primitive sets, while the increase in complexity is only marginal.
Very few (2-3) iterations are needed, as \textbf{uFC} converges very fast.
Increasing the number of iterations has very little impact on $OI$, but results in very complex vocabularies (large $C_0$ and feature lengths).
In the feature set created by \textbf{uFC}$(0.180, 5)$ on \texttt{street}, on average, each feature contains more than 18 literals.
This is obviously too much for human comprehension.

For \textbf{uFRINGE}, the $OI$ indicator shows very marginal or no improvement on \texttt{spect} and \texttt{hungarian} datasets, and even a degradation on \texttt{street} (compared to the primitive set).
Features constructed using this approach have an average length between 2.08 and 3.1 literals, just as much as the selected \textbf{uFC*} configuration.
But, it constructs between 2.6 and 13.9 times more features than \textbf{uFC*}.
We consider this to be due to the lack of filtering in \textbf{uFRINGE}, which would also explain the low $OI$ score.
Old features remain in the feature set and amplify the total correlation by adding the correlation between old and new features.

\subsection{Impact of parameters $\lambda$ and $limit_{iter}$}
\label{subsec:pareto-evaluation}

In order to understand the impact of parameters, we executed \textbf{uFC} with a wide range of values for $\lambda$ and $limit_{iter}$ and studied the evolution of the indicators $OI$ and $C_0$.
For each dataset, we varied $\lambda$ between $0.002$ and $0.5$ with a step of $0.002$.
For each value of $\lambda$, we executed \textbf{uFC} by varying $limit_{iter}$ between $1$ and $30$ for the \texttt{hungarian} dataset, and between $1$ and $20$ for \texttt{street} and \texttt{spect}.
We study the evolution of the indicators as a function of $limit_{iter}$, respectively $\lambda$, we plot the solution in the ($OI, C_0$) space and construct the Pareto front.

\begin{figure}[h]
  \centering
  \subfloat[]{
  	\label{sub-fig:OI-iter}
  	\includegraphics[width=0.48\textwidth]{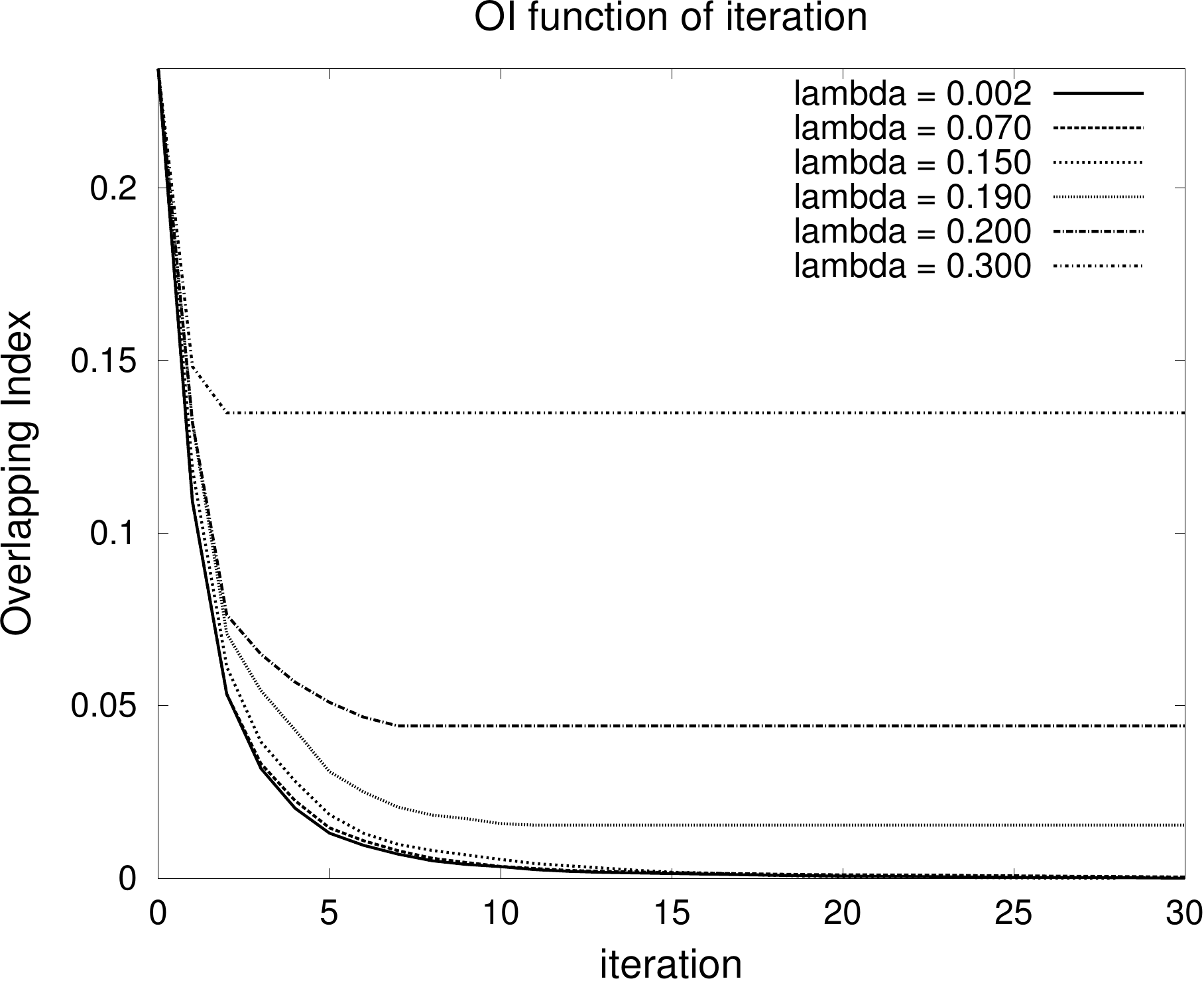}
  }                
  \hfill
  \subfloat[]{
  	\label{sub-fig:OI-lambda}
	\includegraphics[width=0.48\textwidth]{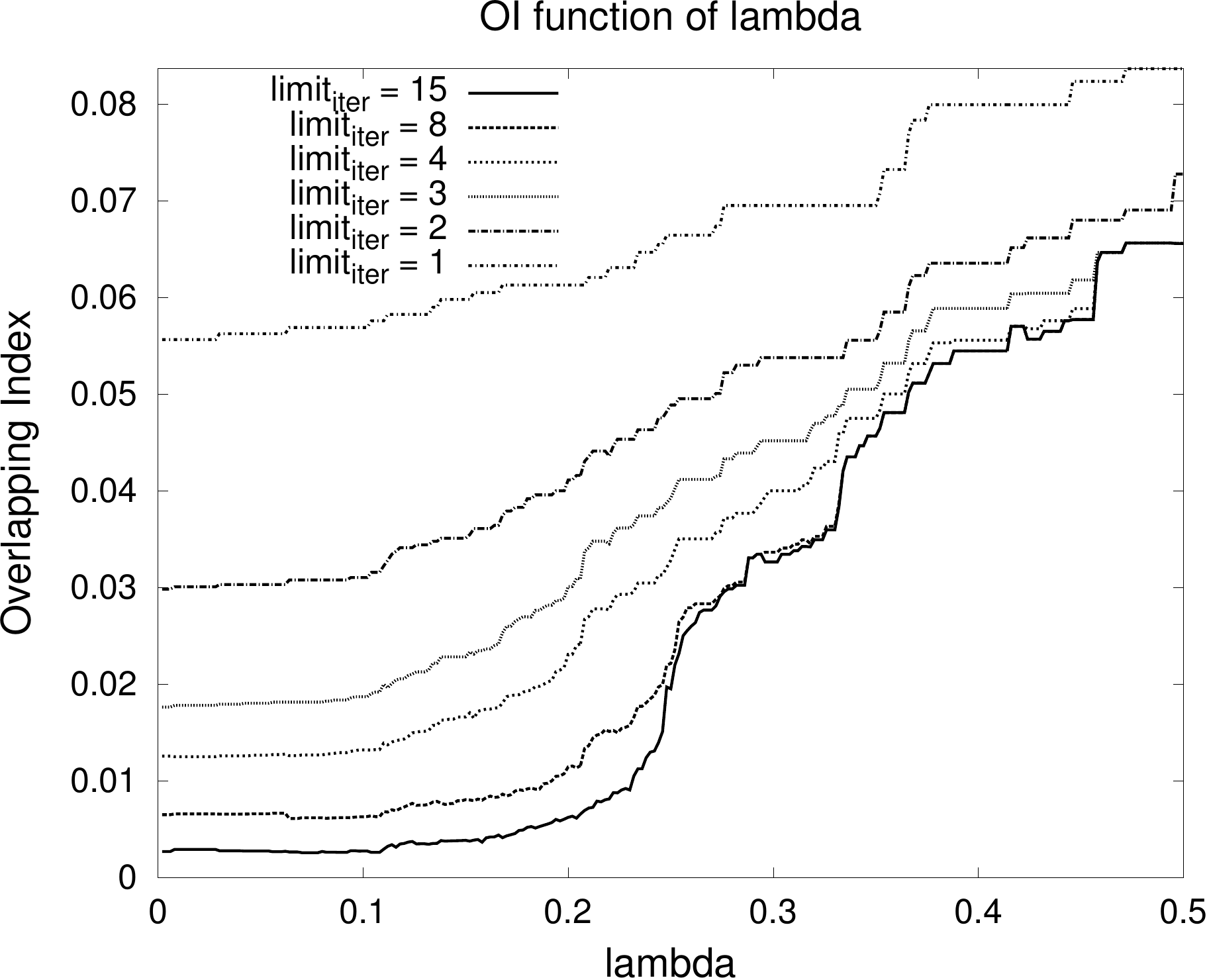}
  }

\caption{Variation of $OI$ indicator with $limit_{iter}$ on \texttt{hungarian} (a) and with $\lambda$ on \texttt{street} (b) }
\label{fig:vary-iteration}
\end{figure}

For the study of $limit_{iter}$, we hold $\lambda$ fixed at various values and we vary only $limit_{iter}$.
The evolution of the $OI$ correlation indicator is given in Fig.~\ref{sub-fig:OI-iter}.
As expected, the measure ameliorates with the number of iterations.
$OI$ has a very rapid descent and needs less than 10 iterations to converge on all datasets towards a value dependent on $\lambda$.
The higher the value of $\lambda$, the higher the value of convergence.
The complexity has a very similar evolution, but in the inverse direction: it increases with the number of iterations performed.
It also converges towards a value that is dependent on $\lambda$: the higher the value of $\lambda$, the lower the complexity of the resulting feature set.

Similarly, we study $\lambda$ by fixing $limit_{iter}$.
Fig.~\ref{sub-fig:OI-lambda} shows how $OI$ evolves when varying $\lambda$.
As foreseen, for all values of $limit_{iter}$, the $OI$ indicator increases with $\lambda$, while $C_0$ decreases with $\lambda$.
$OI$ shows an abrupt increase between 0.2 and 0.3, for all datasets.
For lower values of $\lambda$, many pairs get combined as their correlation score is bigger than the threshold.
As $\lambda$ increases, only highly correlated pairs get selected and this usually happens in the first iterations.
Performing more iterations does not bring any change and indicators are less dependent on $limit_{iter}$.
For \texttt{hungarian}, no pair has a correlation score higher than 0.4.
Setting $\lambda$ higher than this value causes \textbf{uFC} to output the primitive set (no features are created).

\begin{figure}[h]
  \centering
  \subfloat[]{\label{sub-fig:pareto-front}\includegraphics[width=0.518\textwidth]{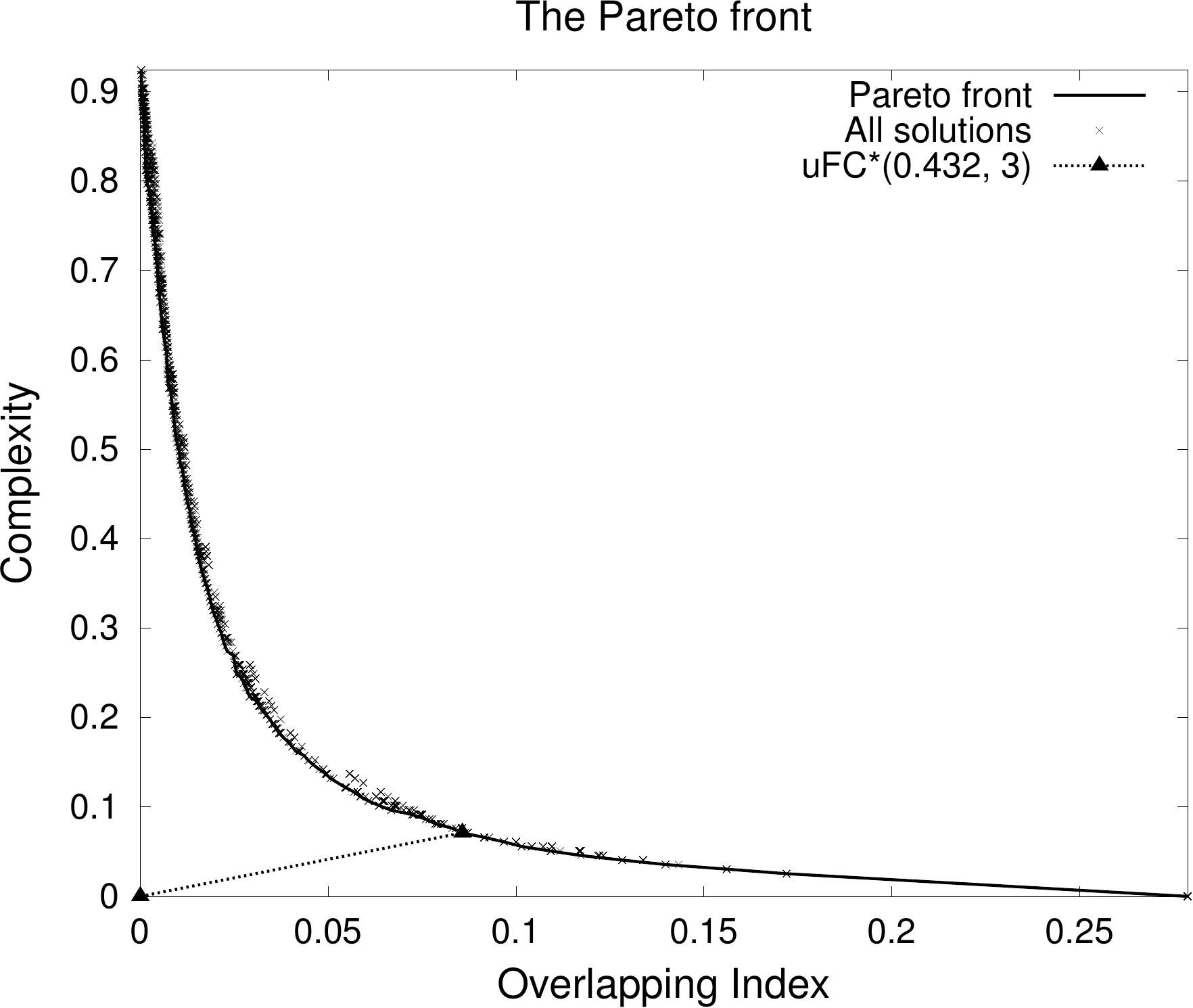}}                
  \hfill
  \subfloat[]{\label{sub-fig:pareto-front-magnified}\includegraphics[width=0.442\textwidth]{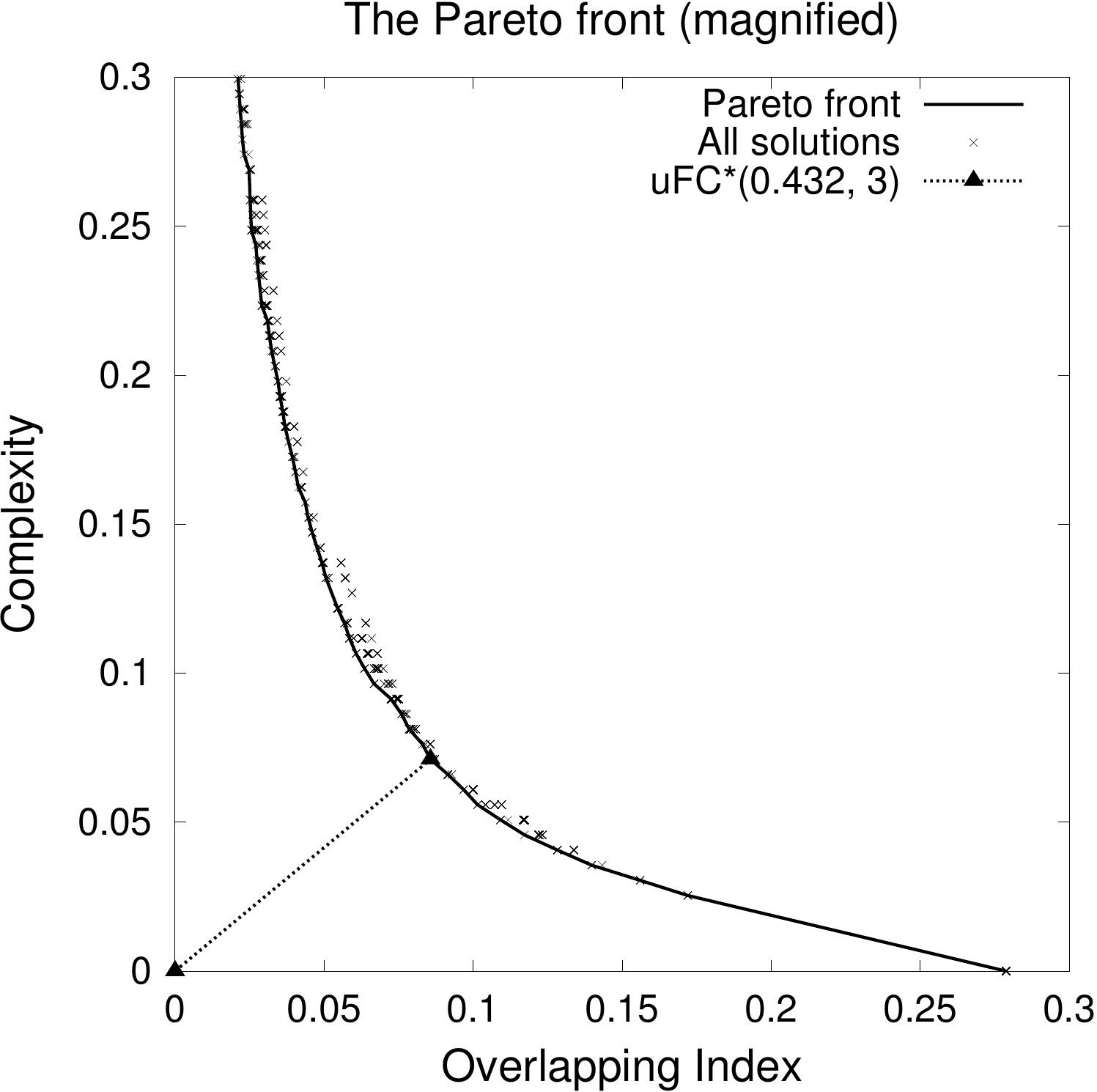}}

\caption{The distribution of solutions, the Pareto front and the closest-point on \texttt{spect} dataset}
\label{fig:pareto-front}
\end{figure}

To study Pareto optimality, we plot the generated solutions in the ($OI$, $C_0$) space.
Fig.~\ref{sub-fig:pareto-front} presents the distribution of solutions, the Pareto front and the solution chosen by the ``closest-point'' heuristic.
The solutions generated by \textbf{uFC} with a wide range of parameter values are not dispersed in the solution space, but their distribution is rather close together.
\textcolor{black}{This shows good algorithm stability.}
Even if not all the solutions are Pareto optimal, none of them are too distant from the front and there are no outliers.

Most of the solutions densely populate the part of the curve corresponding to low $OI$ and high $C_0$.
\textcolor{black}{As pointed out in the Section~\ref{subsec:pareto-front}, the area of the front corresponding to high feature set complexity (high $C_0$) represents the overfitting area.}
This confirms that the algorithm converges fast\textcolor{black}{, then enters overfitting}.
Most of the improvement in quality is done in the first 2-3 iterations, while further iterating improves quality only marginally with the cost of an explosion of complexity.
\textcolor{black}{The ``closest-point'' heuristic keeps the constructing out of overfitting, by stopping the algorithm at the point where the gain of co-occurence score and the loss in complexity are fairly equal.}
Fig.~\ref{sub-fig:pareto-front-magnified} magnifies the region of the solution space corresponding for low numbers of iterations.

\subsection{Relation between number of features and feature length}
\label{subsec:choose-complexity-measure}

Both the average length of a feature ($C_1$) and the number of features ($C_0$) increase with the number of iterations.
In Section~\ref{subsec:complexity-tradeoff} we have speculated that the two are correlated: $C_1 = f(C_0)$.
For each $\lambda$ in the batch of tests, we create the $C_0$ and $C_1$ series depending on the $limit_{iter}$ and we perform a statistical hypothesis test, using the Kendall rank coefficient as the test statistic.
The Kendall rank coefficient is particularly useful as it makes no assumptions about the distributions of $C_0$ and $C_1$.
For all values of $\lambda$, for all datasets, the statistical test revealed a p-value of the order of $10^{-9}$.
This is consistently lower than habitually used significance levels and makes us reject the null independence hypothesis and conclude that $C_0$ and $C_1$ are statistically dependent.

\section{Improving the \textbf{uFC} algorithm}
\label{sec:improvements}

The major difficulty of \textbf{uFC}, shown by the initial experiments, is setting the values of parameters.
An unfortunate choice would result in either an overly complex feature set or a feature set where features are still correlated.
But both parameters $\lambda$ and $limit_{iter}$ are dependent on the dataset and finding the suitable values would prove to be a process of trial and error for each new corpus.
The ``closest-point'' heuristic achieves acceptable equilibrium between complexity and performance, but requires multiple executions with large choices of values for parameters and the construction of the Pareto front, which might not always be desirable or even possible.

We propose a new method for choosing $\lambda$ based on statistical hypothesis testing and a new stopping criterion inspired from the ``closest-point'' heuristic.
These will be integrated into a new ``risk-based'' heuristic that approximates the best solution while avoiding the time consuming construction of multiple solutions and the Pareto front.
The only parameter is the significance level $\alpha$, which is independent of the dataset, and makes the task of running \textbf{uFC} on new, unseen datasets easy.
A pruning technique is also proposed.

\subsection{Automatic choice of $\lambda$}
\label{subsec:choose-lambda}

We propose replacing the user-supplied co-occurrence threshold $\lambda$ with a technique that selects only pairs of features for whom the positive linear correlation is statistically significant.
These pairs are added to the set $O$ of co-occurring pairs (defined in Section~\ref{subsec:new-feat-search}) and, starting from $O$, new features are constructed.
We use a statistical method: the \textit{hypothesis testing}.
For each pair of candidate features, we test the independence hypothesis $H_0$ against the positive correlation hypothesis $H_1$.

We use as a test statistic the Pearson correlation coefficient (calculated as defined in Section~\ref{subsec:new-feat-search}) and test the following formally defined hypothesis: $H_0: \rho=0$ and $H_1: \rho>0$, where $\rho$ is the theoretical correlation coefficient between two candidate features.
We can show that in the case of Boolean variables, having the contingency table shown in Table~\ref{tab:contingency-Table}, the observed value of the $\chi^2$ of independence is $ \chi^2_{obs} = nr^2$ ($n$ is the size of the dataset).
Consequently, considering true the hypothesis $H_0$, $nr^2$ is approximately following a $\chi^2$ distribution with one degree of freedom ($nr^2 \sim \chi^2_1$), resulting in $r\sqrt{n}$ following a standard normal distribution ($r\sqrt{n} \sim N(0,1)$), given that $n$ is large enough.

We reject the $H_0$ hypothesis in favour of $H_1$ if and only if $r\sqrt{n} \geq u_{1-\alpha}$, where $u_{1-\alpha}$ is the right critical value for the standard normal distribution.
Two features will be considered significantly correlated when $ r(\{f_i, f_j\}) \geq  \frac{u_{1-\alpha}}{\sqrt{n}}$.
The significance level $\alpha$ represents the risk of rejecting the independence hypothesis when it was in fact true.
It can be interpreted as the \textit{false discovery risk} in data mining.
In the context of feature construction it is the \textit{false construction risk}, since this is the risk of constructing new features based on a pair of features that are not really correlated.
Statistical literature usually sets $\alpha$ at $0.05$ or $0.01$, but levels of $0.001$ or even $0.0001$ are often used.

The proposed method repeats the independence test a great number of times, which inflates the number of type I errors.
\citet{GE03} presents several methods for controlling the false discoveries.
Setting aside the Bonferroni correction, often considered too simplistic and too drastic, one has the option of using sequential rejection methods~\citep{BEN99,HOL79}, the q-value method of Storey~\citep{STO02} or making use of bootstrap~\citep{LAL06}.
In our case, applying these methods is not clear-cut, as tests performed at each iteration depend on the results of the tests performed at previous iterations.
It is noteworthy that a trade-off must be acquired between the inflation of false discoveries and the inflation of missed discoveries.
This makes us choose a risk between $5\%$ and $\frac{5\%}{m}$, where $m$ is the theoretical number of tests to be performed.

\subsection{Candidate pruning technique. Stopping criterion.}
\label{subsec:candidate-pruning}

\textbf{Pruning} In order to apply the $\chi^2$ independence test, it is necessary that the expected frequencies considering true the $H_0$ hypothesis be greater or equal than 5.
We add this constraint to the new feature search strategy (subsection~\ref{subsec:new-feat-search}).
Pairs for whom the values of $\frac{(a+b)(a+c)}{n}$, $\frac{(a+b)(b+c)}{n}$, $\frac{(a+c)(c+d)}{n}$ and $\frac{(b+d)(c+d)}{n}$ are not greater than 5, will be filtered from the set of candidate pairs $O$.
This will impede the algorithm from constructing features that are present for very few individuals in the dataset.

\textbf{Risk-based heuristic} We introduced in Section~\ref{subsec:pareto-front} the ``closest-point'' for choosing the values for parameters $\lambda$ and $limit_{iter}$.
It searches the solution on the Pareto front for which the indicators are sensibly equal.
We transform the heuristic into a stopping criterion: $OI$ and $C_0$ are combined into a single formula, the \textbf{root mean square} (RMS).
The algorithm will stop iterating when RMS has reached a minimum.
Using the generalized mean inequality, we can prove that $RMS(OI, C_0)$ has only one global minimum, as with each iteration the complexity increases and $OI$ descends.

The $limit_{iter}$ parameter, which is data-dependent, is replaced by the automatic \textit{RMS} stopping criterion.
This stopping criterion together with the automatic $\lambda$ choice strategy, presented in Section~\ref{subsec:choose-lambda}, form a data-independent heuristic for choosing parameters.
We will call the new heuristic \textbf{risk-based heuristic}.
This new heuristic will make it possible to approximate the best parameter compromise and avoid the time consuming task of computing a batch of solutions and constructing the Pareto front.

\section{Further Experiments}
\label{sec:second-xp}

We test the proposed ameliorations, similarly to what was shown in Section~\ref{sec:xps}, on the same three datasets: \texttt{hungarian}, \texttt{spect} and \texttt{street}.
We execute \textbf{uFC} in two ways: the classical \textbf{uFC} (Section~\ref{sec:our-proposition}) and the improved \textbf{uFC} (Section~\ref{sec:improvements}).
The classical \textbf{uFC} needs to have parameters  $\lambda$ and $limit_{iter}$ set (noted \textbf{uFC}$(\lambda$,~$limit_{iter})$).
\textbf{uFC*}$(\lambda$,~$limit_{iter})$ denotes the execution with parameters which were determined \textit{a posteriori} using the ``closest-point'' heuristic.
The improved \textbf{uFC} will be denoted as \textbf{uFC}$_{\alpha}$($risk$).
The ``risk-based'' heuristic will be used to determine the parameters and control the execution.

\subsection{Risk-based heuristic for choosing parameters}

\begin{figure}[!t]
	\centering
	\includegraphics[width=0.518\textwidth]{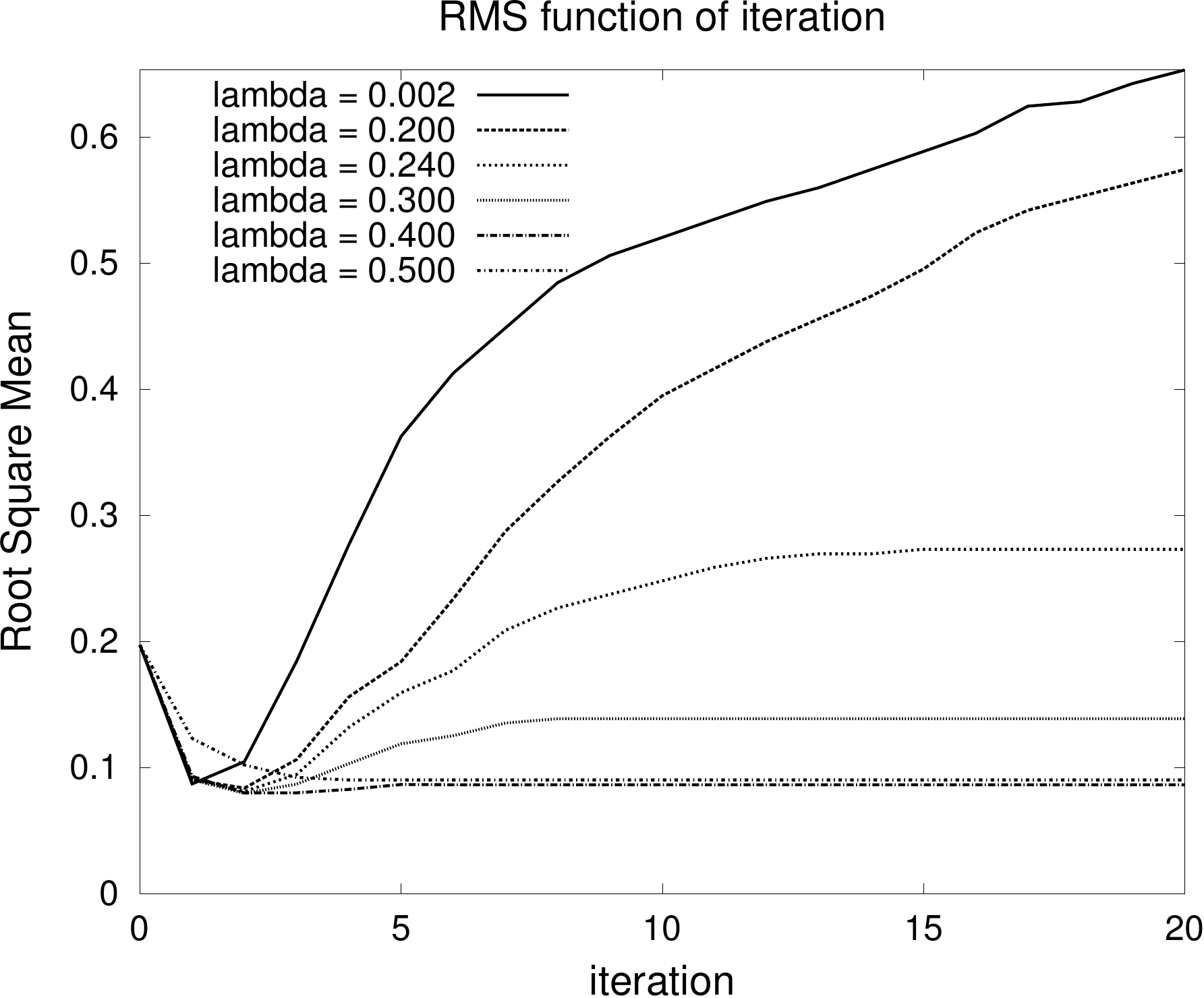}
	\caption{RMS vs. $limit_{iter}$ on \texttt{spect}}
	\label{sub-fig:spect-rms}
\end{figure}

\textbf{Root Mean Square} In the first batch of experiments, we study the variation of the Root Means Square aggregation function for a series of selected values of $\lambda$.
We vary $limit_{iter}$ between 0 and 30, for \texttt{hungarian}, and between 0 and 20 for \texttt{spect} and \texttt{street}.
The evolution of RMS is presented in Fig.~\ref{sub-fig:spect-rms}.

For all $\lambda$ the RMS starts by decreasing, as $OI$ descends more rapidly than the $C_0$ increases.
In just 1-3 iterations, RMS reaches its minimum and afterwards its value starts to increase.
This is due to the fact that complexity increases rapidly, with only marginal improvement of quality.
This behaviour is consistent with the results presented in Section~\ref{sec:xps}.
As already discussed in Section~\ref{subsec:pareto-evaluation}, $\lambda$ has a bounding effect over complexity, thus explaining why RMS reaches a maximum for higher values of $\lambda$.

\begin{figure}[!t]
	\centering
	\includegraphics[width=0.442\textwidth]{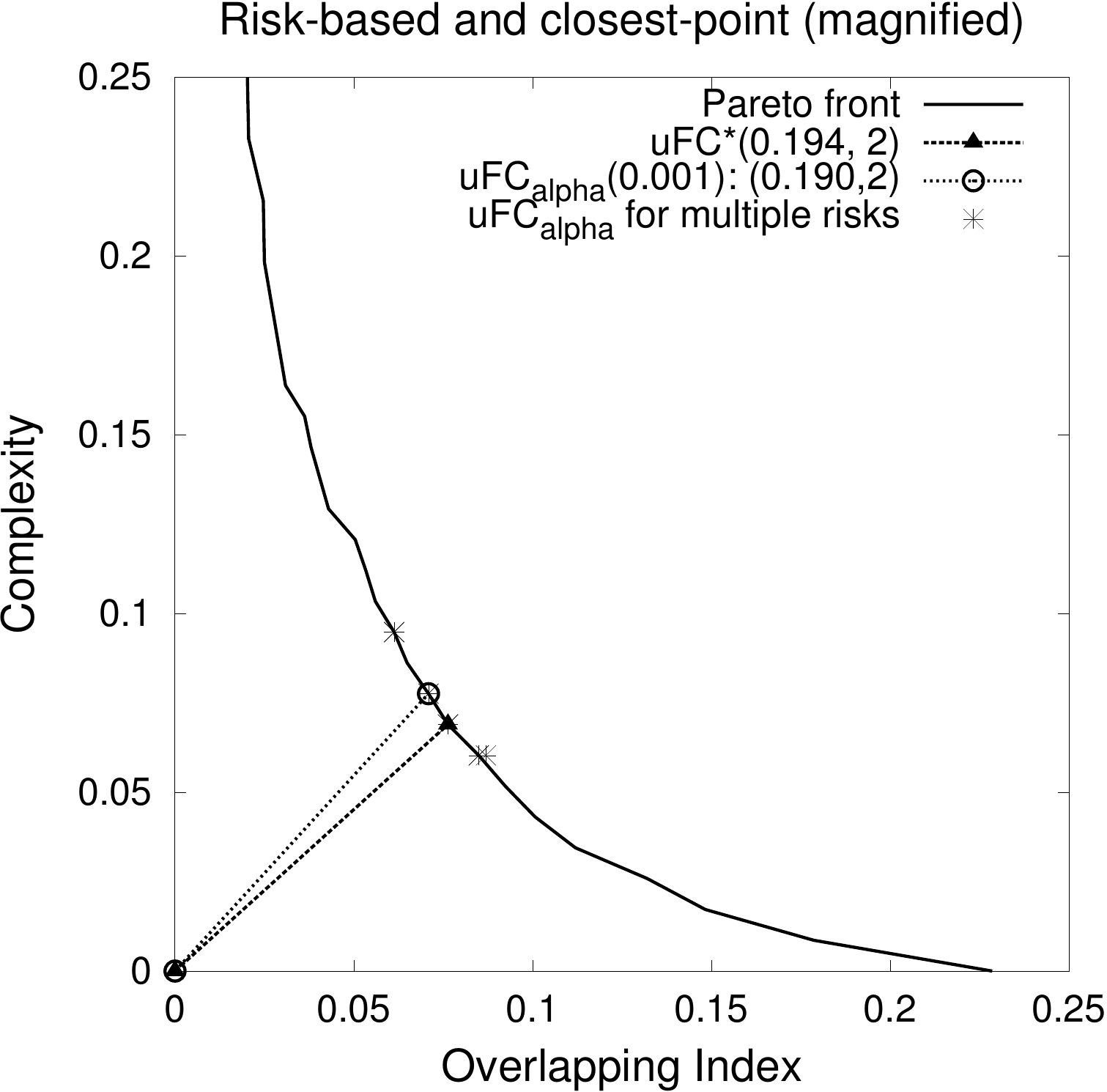}
	\caption{``Closest-point'' and ``risk-based'' for multiple $\alpha$ on \texttt{hungarian}}
	\label{sub-fig:hu-risk}
\end{figure}

\textbf{The ``risk-based'' heuristic} The second batch of experiments deals with comparing the ``risk-based'' heuristic to the ``closest-point'' heuristic.
The ``closest-point'' was determined as described in Section~\ref{sec:xps}.
The ``risk-based'' heuristic was executed multiple times, with values for parameter $\alpha \in \{ 0.05$, $0.01$, $0.005$, $0.001$, $0.0008$, $0.0005$, $0.0003$, $0.0001$, $0.00005$, $0.00001 \}$

\begin{table*}[htbp]
\caption{``closest-point'' and ``risk-based'' heuristics}
\label{tab:result-risk-based}

\centering
\begin{tabular}{llccccccc}
\toprule
 & \textbf{Strategy} & \textbf{$\lambda$} & \textbf{$limit_{iter}$} &\textbf{$\#feat$} & \textbf{$\#common$} & \textbf{$length$} & \textbf{$OI$} & \textbf{$C_0$} \\ \midrule
 
\multirow{3}{*}{\begin{sideways}\texttt{hung.}\end{sideways}} & \textbf{Primitives} & - & - & 13 & - & 1.00 & 0.235 & 0.000 \\  
& \textbf{uFC*(0.194, 2)} & 0.194 & 2 & 21 & \multirow{2}{*}{19} & 2.95 & 0.076 & 0.069 \\ 
& \textbf{uFC$_{\alpha}$(0.001)} & 0.190 & 2 & 22 & & 3.18 & 0.071 & 0.078 \\ \cmidrule{2-9}

\multirow{3}{*}{\begin{sideways}\texttt{street}\end{sideways}} & \textbf{Primitives} & - & - & 66 & - & 1.00 & 0.121 & 0.000 \\ 
& \textbf{uFC*(0.446, 3)} & 0.446 & 3 & 87 & \multirow{2}{*}{33} & 2.14 & 0.062 & 0.038 \\ 
& \textbf{uFC$_{\alpha}$(0.0001)} & 0.150 & 1 & 90 & & 1.84 & 0.060 & 0.060 \\ \cmidrule{2-9}

\multirow{3}{*}{\begin{sideways}\texttt{spect}\end{sideways}} & \textbf{Primitives} & - & - & 22 & - & 1.00 & 0.279 & 0.000 \\
& \textbf{uFC*(0.432, 3)} & 0.432 & 3 & 36 & \multirow{2}{*}{19} & 2.83 & 0.086 & 0.071 \\ 
& \textbf{uFC$_{\alpha}$(0.0001)} & 0.228 & 2 & 39 & & 2.97 & 0.078 & 0.086 \\ \bottomrule
\end{tabular}
\end{table*}

Table~\ref{tab:result-risk-based} gives a quantitative comparison between the two heuristics.
A risk of $0.001$ is used for \texttt{hungarian} and $0.0001$ for \texttt{spect} and \texttt{street}.
The feature sets created by the two approaches are very similar, considering all indicators.
Not only the differences between values for $OI$, $C_0$, average feature length and feature set dimension are negligible, but most of the created features are identical.
On \texttt{hungarian}, 19 of the 21 features created by the two heuristics are identical.
Table~\ref{tab:exec-results-risk} shows the two features sets, with non-identical features in bold.

\begin{table*}[htbp]
\caption{Feature sets constructed by ``closest-point'' and ``risk-based'' heuristics on \texttt{hungarian}}
\label{tab:exec-results-risk}

\centering
\begin{tabular}{lll}
\toprule
\multicolumn{1}{c}{\textbf{primitives}} & \multicolumn{1}{c}{\textbf{uFC*(0.194, 2)}} & \multicolumn{1}{c}{\textbf{uFC$_{\alpha}$(0.001)}} \\ \midrule
$person$ & $groups \wedge \overline{road} \wedge interior $ & $ groups \wedge \overline{road} \wedge interior $ \\ 
$groups$ & $groups \wedge \overline{road} \wedge \overline{interior} $ & $ groups \wedge \overline{road} \wedge \overline{interior} $ \\ 
$water$ & $\overline{groups \wedge \overline{road}} \wedge interior $ & $ \overline{groups \wedge \overline{road}} \wedge interior $ \\ 
$cascade$ & $water \wedge cascade \wedge tree \wedge forest $ & $ water \wedge cascade \wedge tree \wedge forest $ \\ 
$ sky$ & $water \wedge cascade \wedge \overline{tree \wedge forest} $ & $ water \wedge cascade \wedge \overline{tree \wedge forest} $ \\ 
$ tree$ & $\overline{water \wedge cascade} \wedge tree \wedge forest $ & $ \overline{water \wedge cascade} \wedge tree \wedge forest $ \\ 
$ grass$ & $sky \wedge building \wedge tree \wedge \overline{forest} $ & $ sky \wedge building \wedge tree \wedge \overline{forest} $ \\ 
$ forest$ & $sky \wedge building \wedge \overline{tree \wedge \overline{forest}} $ & $ sky \wedge building \wedge \overline{tree \wedge \overline{forest}} $ \\ 
$ statue$ & $\overline{sky \wedge building} \wedge tree \wedge \overline{forest} $ & $ \overline{sky \wedge building} \wedge tree \wedge \overline{forest} $ \\ 
$building$ & $sky \wedge \overline{building} \wedge panorama $ & $ sky \wedge \overline{building} \wedge panorama $ \\ 
$ road$ & $sky \wedge \overline{building} \wedge \overline{panorama} $ & $ sky \wedge \overline{building} \wedge \overline{panorama} $ \\ 
$ interior$ & $\overline{sky \wedge \overline{building}} \wedge panorama $ & $ \overline{sky \wedge \overline{building}} \wedge panorama $ \\ 
$panorama$ & $\overline{groups} \wedge road \wedge person $ & $ \overline{groups} \wedge road \wedge person $ \\ 
 & $\overline{groups} \wedge road \wedge \overline{person} $ & $ \overline{groups} \wedge road \wedge \overline{person} $ \\ 
 & $\overline{\overline{groups} \wedge road} \wedge person $ & $ \overline{\overline{groups} \wedge road} \wedge person $ \\ 
 & $water \wedge \overline{cascade} $ & $ \mathbf{\overline{sky} \wedge building \wedge groups \wedge road }$ \\ 
 & $\mathbf{\overline{sky} \wedge building} $ & $ \mathbf{\overline{sky} \wedge building \wedge \overline{groups \wedge road} }$ \\ 
 & $\overline{tree} \wedge forest $ & $ \mathbf{\overline{\overline{sky} \wedge building} \wedge groups \wedge road }$ \\ 
 & $\mathbf{groups \wedge road }$ & $ water \wedge \overline{cascade} $ \\ 
 & $grass $ & $ \overline{tree} \wedge forest $ \\ 
 & $statue $ & $ grass $ \\ 
 &  & $ statue $ \\ \bottomrule
\end{tabular}
\end{table*}

Fig.~\ref{sub-fig:hu-risk} presents the distribution of solutions created by the ``risk-based'' heuristic with multiple $\alpha$, plotted on the same graphics as the Pareto front in the ($OI$, $C_0$) space.
Solutions for different values of risk $\alpha$ are grouped closely together.
Not all of them are on the Pareto front, but they are never too far from the ``closest-point'' solution, providing a good equilibrium between quality and complexity.

\begin{figure}[!h]
  \centering
  \subfloat[]{
  	\label{sub-fig:street-risk-based}
  	\includegraphics[width=0.48\textwidth]{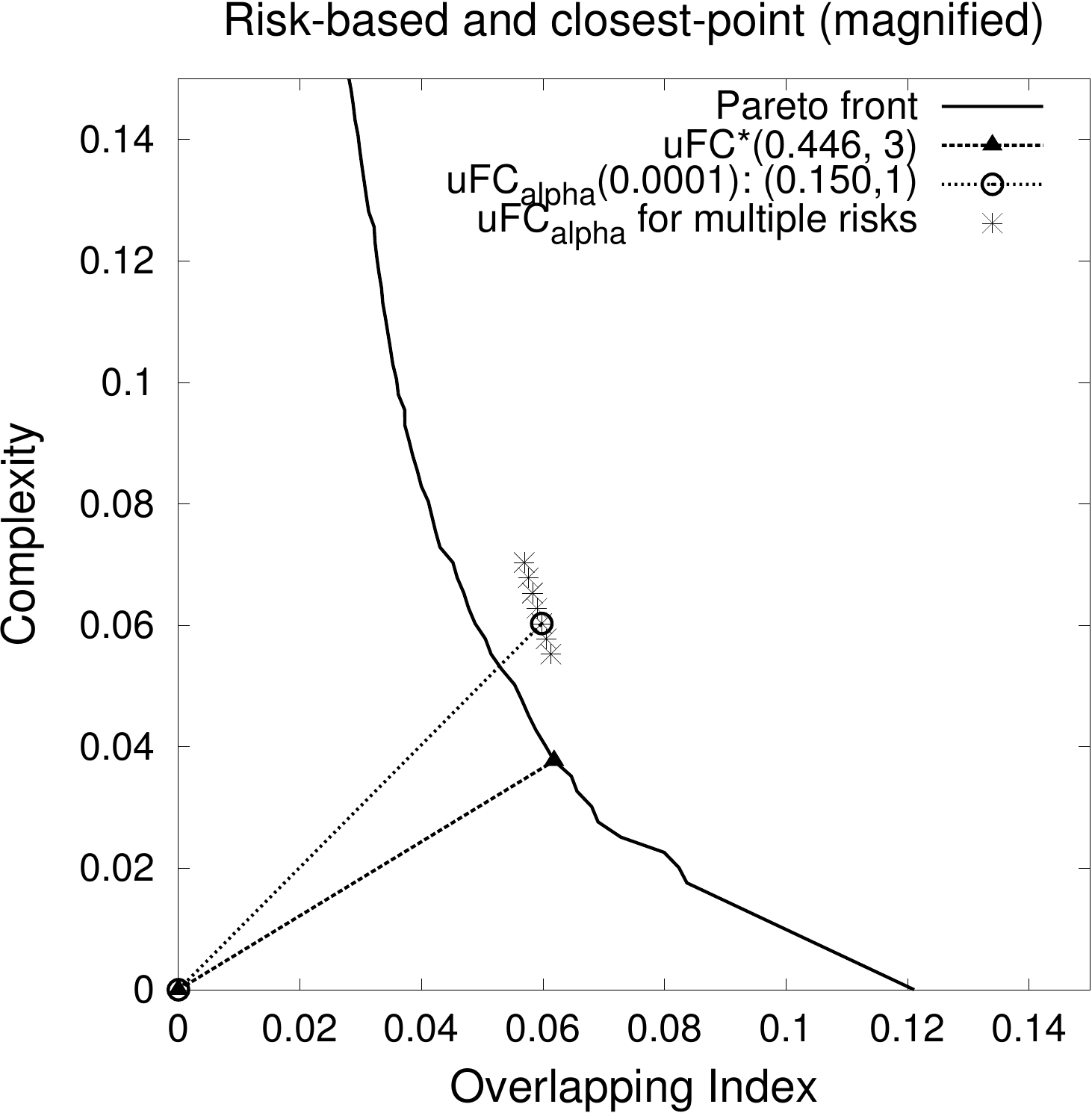}
  }
  \hfill
  \subfloat[]{
  	\label{sub-fig:street-risk-based-F}
  	\includegraphics[width=0.48\textwidth]{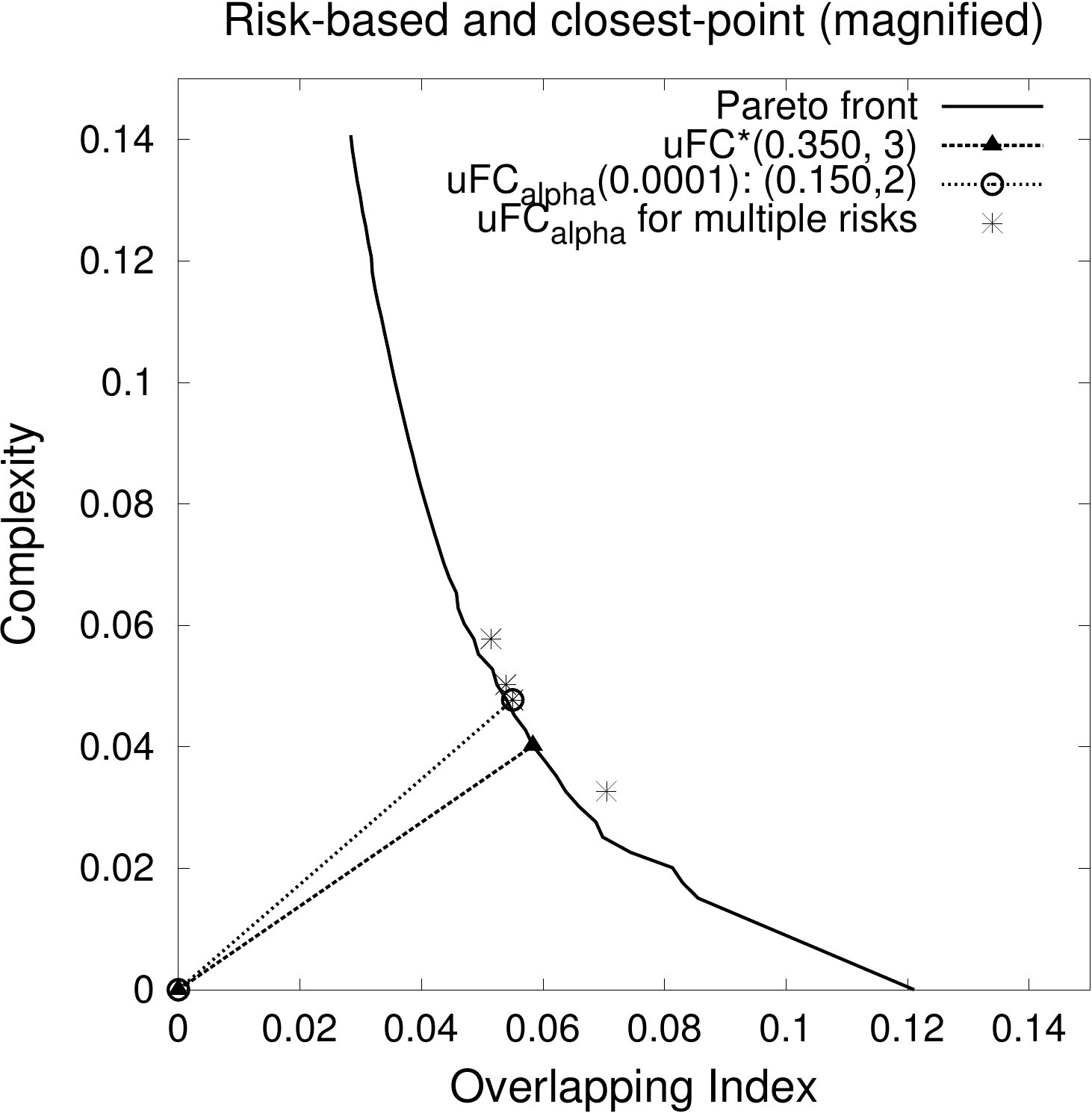}
  }

	\caption{``Closest-point'' and ``Risk-based'' heuristics for \texttt{street} without pruning (a) and with pruning (b)}
	\label{fig:street-risk-based}
\end{figure}

On \texttt{street}, performances of the ``risk-based'' heuristic start to degrade compared to \textbf{uFC*}.
Table~\ref{tab:result-risk-based} shows differences in the resulted complexity and only $33\%$ of the constructed features are common for the two approaches.
Fig.~\ref{sub-fig:street-risk-based} shows that solutions found by the ``risk-based'' approach are moving away from the ``closest-point''.
The cause is the large size of the \texttt{street} dataset.
As the sample size increases, the null hypothesis tends to be rejected at lower levels of p-value.
The auto-determined $\lambda$ threshold is set too low and the constructed feature sets are too complex.
Pruning solves this problem as shown in Fig.~\ref{sub-fig:street-risk-based-F} and Section~\ref{subsec:pruning}.

\subsection{Pruning the candidates}
\label{subsec:pruning}

The pruning technique is independent of the ``risk-based'' heuristic and can be applied in conjunction with the classical \textbf{uFC} algorithm.
An execution of this type will be denoted \textbf{uFC}$_P(\lambda, max_{iter})$.
We execute \textbf{uFC}$_P(\lambda, max_{iter})$ with the same parameters and on the same datasets as described in Section~\ref{subsec:pareto-evaluation}.

\begin{figure}[!h]

  \centering
  
  \subfloat[]{
  	\label{sub-fig:hu-F_and_NF}
  	\includegraphics[width=0.47\textwidth]{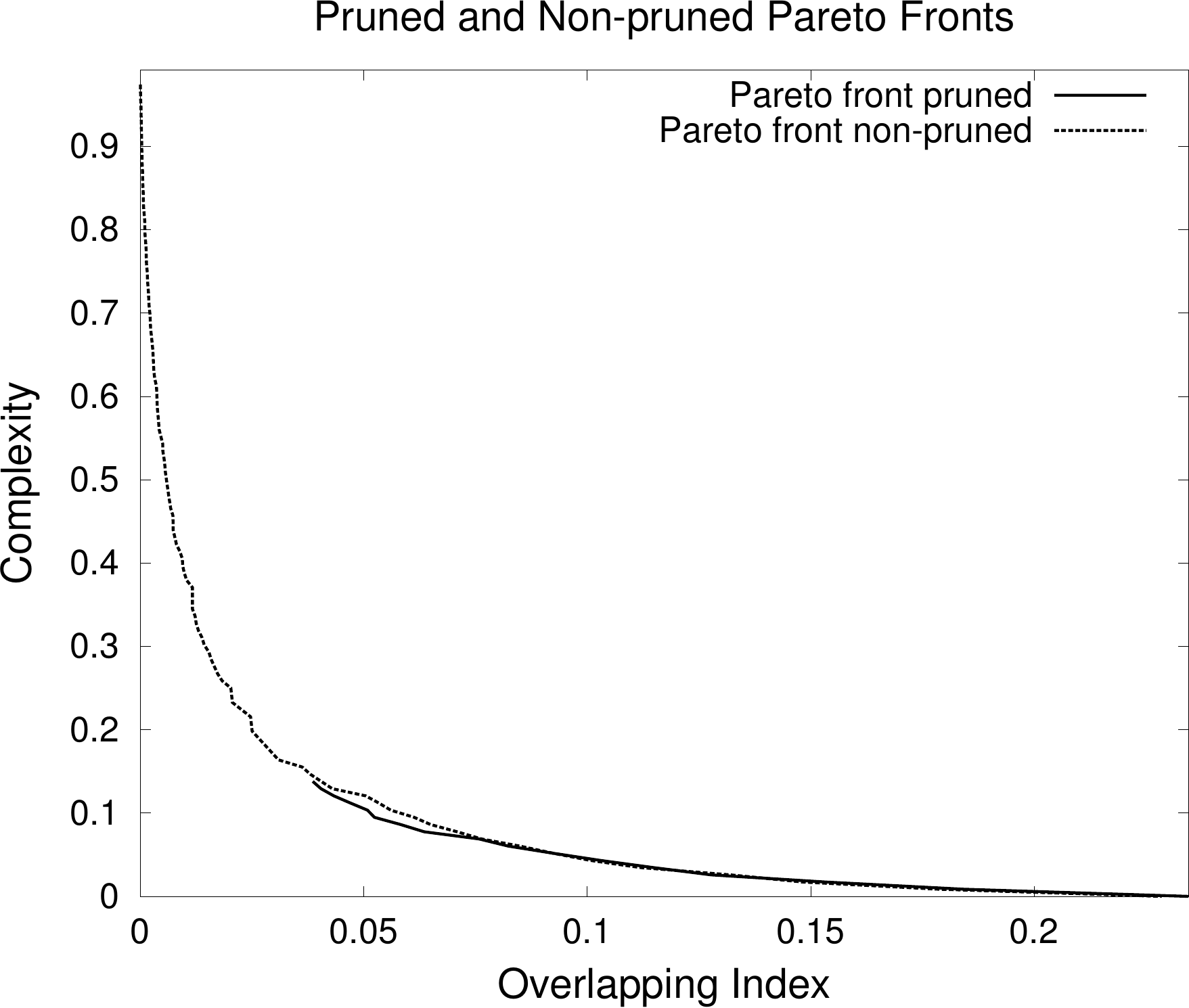}
  }                
  \hfill
  \subfloat[]{
  	\label{sub-fig:hu-F_and_NF_magnify}
  	\includegraphics[width=0.49\textwidth]{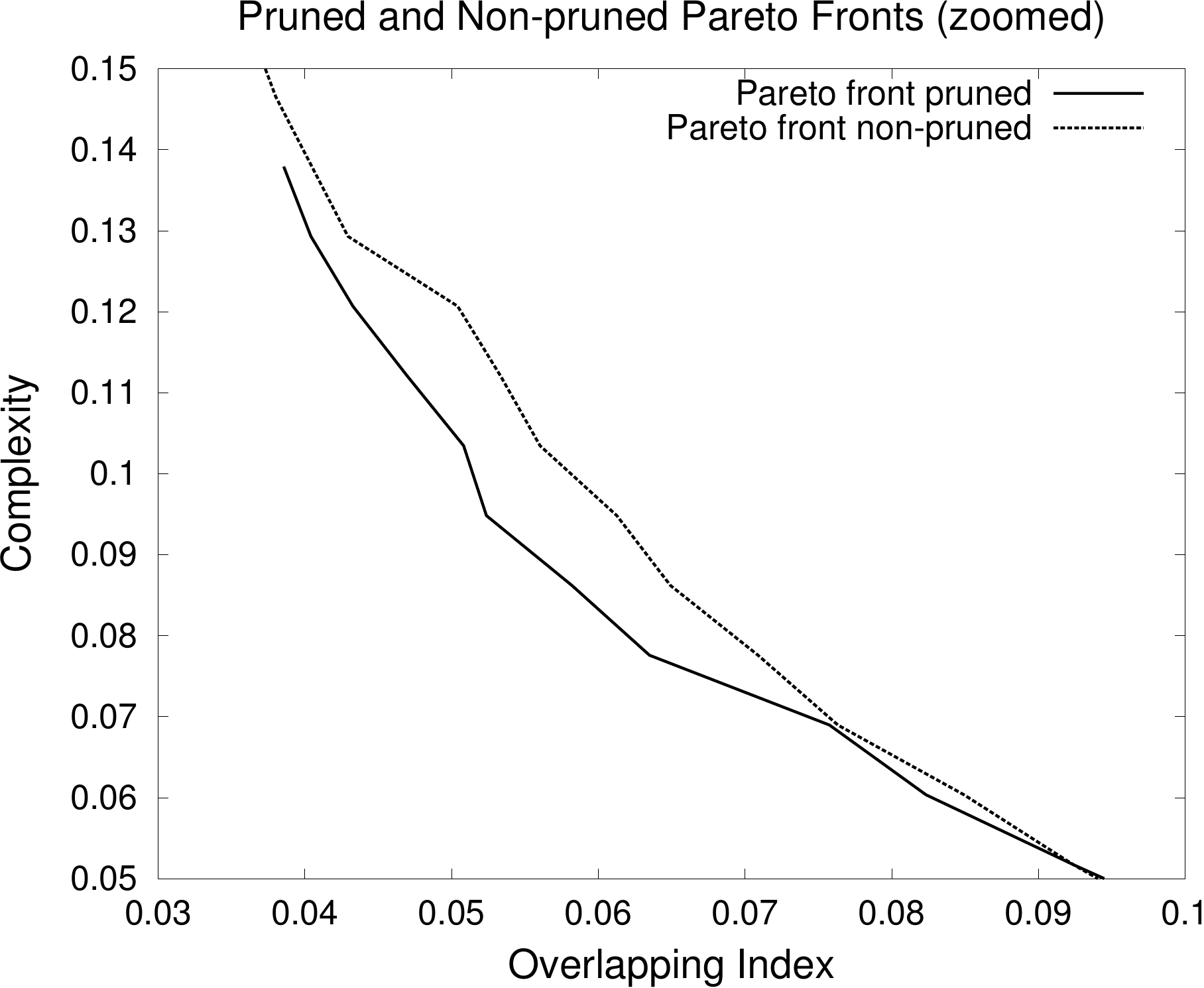}
  }

	\caption{Pruned and Non-pruned Pareto Fronts on \texttt{hungarian} (a) and a zoom to the relevant part (b)}
	\label{fig:F-and-NF-pareto}
\end{figure}

We compare \textbf{uFC} with and without pruning by plotting on the same graphic the two Pareto fronts resulted from each set of executions.
Fig.~\ref{sub-fig:hu-F_and_NF} shows the pruned and non-pruned Pareto fronts on \texttt{hungarian}.
The graphic should be interpreted in a manner similar to a ROC curve, since the algorithm seeks to minimize $OI$ and $C_0$ at the same time.
When one Pareto front runs closer to the origin of the graphic $(0,0)$ than a second, it means that the first dominates the second one and, thus, its corresponding approach yields better results.
For all datasets, the pruned Pareto front dominates the non-pruned one.
The difference is marginal, but proves that filtering improves results.

The most important conclusion is that filtering limits complexity.
As the initial experiments (Fig.~\ref{sub-fig:pareto-front}) showed, most of the non-pruned solutions correspond to very high complexities.
Visually, the Pareto front is tangent to the vertical axis (the complexity) and showing complexities around $0.8-0.9$ (out of 1).
On the other hand, the Pareto front corresponding to the pruned approach stops, for all datasets, for complexities lower than $0.15$.
This proves that filtering successfully discards solutions that are too complex to be interpretable.

Last, but not least, filtering corrects the problem of automatically choosing $\lambda$ for the ``risk-based'' heuristic on big datasets.
We ran \textbf{uFC}$_P$ with risk $\alpha \in \{ 0.05$, $0.0$1, $0.005$, $0.001$, $0.0008$, $0.0005$, $0.0003$, $0.0001$, $0.00005$, $0.00001 \}$.
Fig.~\ref{sub-fig:street-risk-based-F} presents the distributions of solutions found with the ``risk-based pruned'' heuristic on \texttt{street}.
Unlike results without pruning (Fig.~\ref{sub-fig:street-risk-based}), solutions generated with pruning are distributed closely to those generated by ``closest-point'' and to the Pareto front.

\subsection{\textcolor{black}{Algorithm stability}}
\label{subsec:stability}

\textcolor{black}{In order to evaluate the stability of the \textbf{uFC}$_{\alpha}$ algorithm, we introduce noise in the \texttt{hungarian} dataset.
The percentage of noise varied between 0\% (no noise) and 30\%.
Introducing a certain percentage $x\%$ of noise means that $x\% \times k \times n$ random features in the datasets are inverted (false becomes true and true becomes false).
$k$ is the number of primitives and $n$ is the number of individuals.
For each given noise percentage, 10 noised datasets are created and only the averages are presented.
\textbf{uFC}$_{\alpha}$ is executed for all the noised datasets, with the same combination of parameters ($risk=0.001$ and no filtering).}

\begin{figure}[!h]

  \centering
  
  \subfloat[]{
  	\label{sub-fig:stability-indicators}
  	\includegraphics[width=0.48\textwidth]{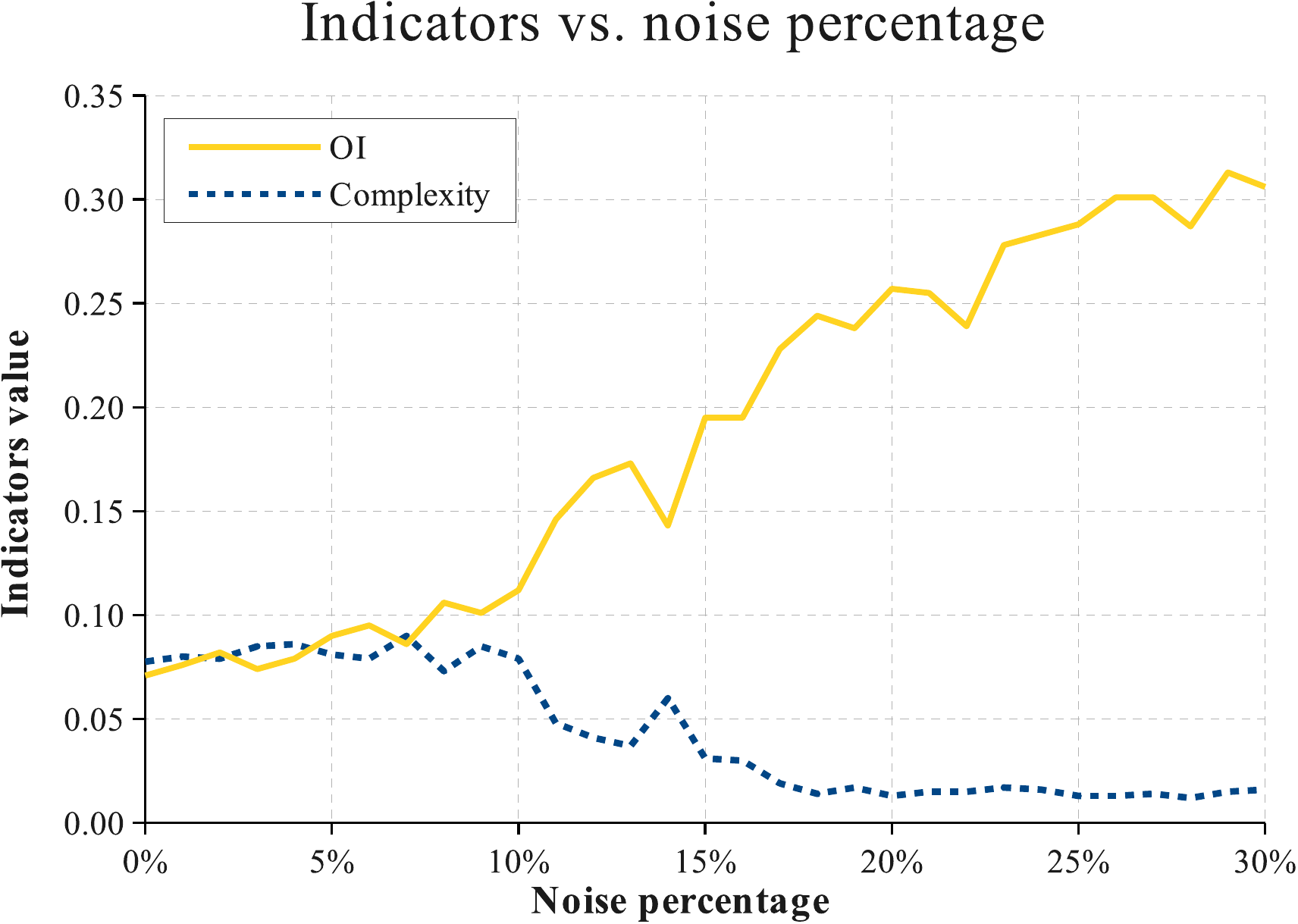}
  }                
  \hfill
  \subfloat[]{
  	\label{sub-fig:stability-no-feat}
  	\includegraphics[width=0.48\textwidth]{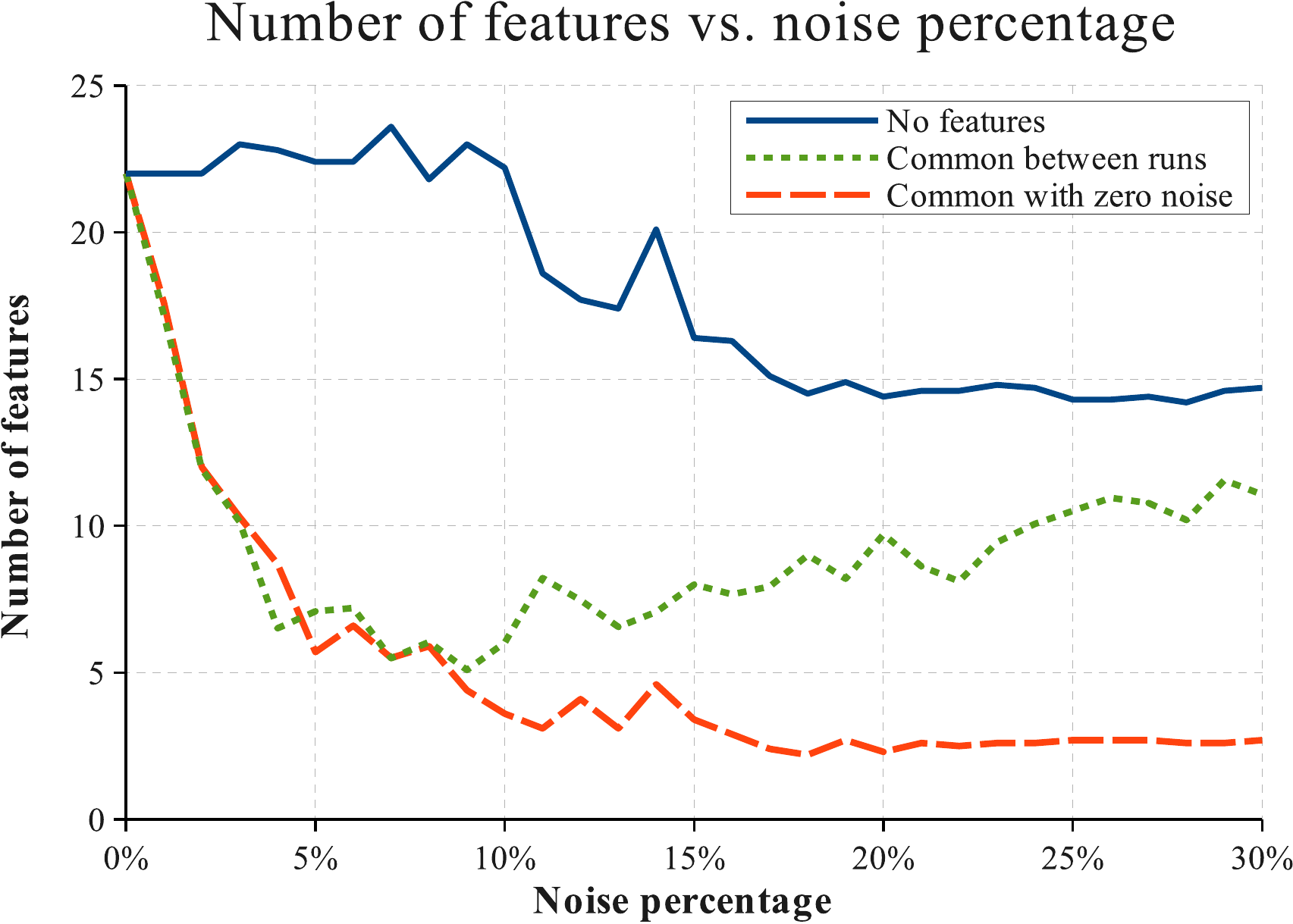}
  }

	\caption{\textcolor{black}{\textbf{uFC}$_{\alpha}$($risk$) stability on \texttt{hungarian} when varying the noise percentage: and indicators (a) and number of constructed features (b)}}
	\label{fig:stability}
\end{figure}

\textcolor{black}{The stability is evaluated using five indicators:
\begin{itemize}
	\item \textbf{Overlapping Index};
	\item \textbf{Feature set complexity} ($C_0$);
	\item \textbf{Number of features}: the total number of features constructed by the algorithm;
	\item \textbf{Common with zero noise}: the number of identical features between the feature sets constructed based on the noised datasets and the non-noised dataset.
	This indicator evaluates the measure in which the algorithm is capable of constructing the same features, even in the presence of noise;
	\item \textbf{Common between runs}: the average number of identical features between feature sets constructed using datasets with the same noise percentage.
	This indicator evaluates how much the constructed feature sets differ at the same noise level.
\end{itemize}
}

\textcolor{black}{As the noise percentage augments, the dataset becomes more random. 
Less pairs of primitives are considered as correlated and therefore less new features are created.
Fig.~\ref{sub-fig:stability-indicators} shows that the overlapping indicator increases with the noise percentage, while the complexity decreases.
Furthermore, most features in the initial dataset are set to false.
As the percentage of noise increases, the ratio equilibrates (more false values becoming true, than the contrary).
As a consequence, for high noise percentages, the OI score is higher than for the primitive set.}

\textcolor{black}{
The same conclusions can be drawn from Fig.~\ref{sub-fig:stability-no-feat}.
The indicator \textbf{Number of features} descends when the noise percentage increases.
This is because fewer features are constructed and the resulting feature set is very similar to the primitive set.
The number of constructed features stabilizes around 20\% of noise.
This is the point where most of the initial correlation between features is lost.
\textbf{Common with zero noise} has a similar evolution.
The number of features identical to the non-noised dataset descends quickly and stabilizes around 20\%.
After 20\%, all the identical features are among the initial primitives.
Similarly, the value of \textbf{Common between runs} descends at first.
For small values of introduced noise, the correlation between certain features is reduced, modifying the order in which pairs of correlated features are selected in Algorithm~\ref{algo:our-proposition}.
This results in a diversity of constructed feature sets.
As the noise level increases and the noised datasets become more random, the constructed feature sets resemble the primitive set, therefore augmenting the value of \textbf{Common between runs}.
}

\section{Conclusion and future work}
\label{sec:conclusion-future-work}

In this article, we propose two approaches towards feature construction.
Unlike the other feature construction algorithms proposed so far in the literature, our proposals work in an unsupervised learning paradigm.
\textbf{uFRINGE} is an unsupervised adaptation of the FRINGE algorithm, while \textbf{uFC} is a new approach that replaces linearly correlated features with conjunctions of literals.
We prove that our approaches succeed in reducing the overall correlation in the feature set, while constructing comprehensible and interpretable features.
We have performed extensive experiments to highlight the impact of parameters on the total correlation measure and feature set complexity.
Based on the first set of experiments, we have proposed a heuristic that finds a suitable balance between quality and complexity and avoids time consuming multiple executions, followed by a Pareto front construction.
We use statistical hypothesis testing and confidence levels for parameter approximation and reasoning on the Pareto front of the solutions for evaluation.
We also propose a pruning technique, based on hypothesis testing, that limits the complexity of the generated features and speeds up the construction process.

For future development, we consider taking into account non-linear correlation between variables by modifying the metric of the search and the co-occurence measure.
Another research direction will be adapting our algorithms for data of the Web 2.0 (e.g., automatic treatment of labels on the web).
Several challenges arise, like very large label sets (it is common to have over 10 000 features), non-standard label names (see standardization preprocessing task that we have performed for the LabelMe dataset in Section~\ref{sec:xps}) and missing data (a value of \textbf{false} can mean absence or missing data).
We also consider converting generated features to the Disjunctive Normal Form for easier reading and suppressing features that have a low support in the dataset.
This would reduce the size of the feature set by removing rare features, but would introduce new difficulties such as detecting nuggets.



\end{document}